\documentclass[letterpaper]{article} % DO NOT CHANGE THIS
\usepackage{aaai25}  % DO NOT CHANGE THIS
\usepackage{times}  % DO NOT CHANGE THIS
\usepackage{helvet}  % DO NOT CHANGE THIS
\usepackage{courier}  % DO NOT CHANGE THIS
\usepackage[hyphens]{url}  % DO NOT CHANGE THIS
\usepackage{graphicx} % DO NOT CHANGE THIS
\usepackage{natbib}  % DO NOT CHANGE THIS AND DO NOT ADD ANY OPTIONS TO IT
\usepackage{caption} % DO NOT CHANGE THIS AND DO NOT ADD ANY OPTIONS TO IT
\usepackage{amsmath}
\usepackage{amssymb}
\usepackage{algorithm}
\usepackage{algorithmic}
\usepackage{multirow}
\usepackage{graphicx}
\usepackage{colortbl}
\usepackage{booktabs}
\usepackage{newfloat}
\usepackage{listings}
\usepackage{xcolor}

\definecolor{coral}{RGB}{255,127,80}
\newcommand\hl[1]{\textcolor{coral}{#1}}
\DeclareCaptionStyle{ruled}{labelfont=normalfont,labelsep=colon,strut=off} % DO NOT CHANGE THIS
\floatstyle{ruled}
\newfloat{listing}{tb}{lst}{}
\floatname{listing}{Listing}
\title{MMAD-Purify: A Precision-Optimized Framework for Efficient and Scalable Multi-Modal Attacks}
\author {
    Xinxin Liu\textsuperscript{\rm 1},
    Zhongliang Guo\textsuperscript{\rm 2},
    Siyuan Huang\textsuperscript{\rm 1},
    Chun Pong Lau\textsuperscript{\rm 3}
}
\affiliations {
    \textsuperscript{\rm 1}John Hopkins University\\
    \textsuperscript{\rm 2}University of St Andrews\\
    \textsuperscript{\rm 3}City University of Hong Kong\\
    \{xliu197,shuan124\}@jh.edu,
    zg34@st-andrews.ac.uk,
    cplau27@cityu.edu.hk
}

\begin{document}

\maketitle

\begin{abstract}
%\vspace{15em}

Neural networks have achieved remarkable performance across a wide range of tasks, yet they remain susceptible to adversarial perturbations, which pose significant risks in safety-critical applications. With the rise of multimodality, diffusion models have emerged as powerful tools not only for generative tasks but also for various applications such as image editing, inpainting, and super-resolution. However, these models still lack robustness due to limited research on attacking them to enhance their resilience. Traditional attack techniques, such as gradient-based adversarial attacks and diffusion model-based methods, are hindered by computational inefficiencies and scalability issues due to their iterative nature. To address these challenges, we introduce an innovative framework that leverages the distilled backbone of diffusion models and incorporates a precision-optimized noise predictor to enhance the effectiveness of our attack framework. This approach not only enhances the attack's potency but also significantly reduces computational costs. Our framework provides a cutting-edge solution for multi-modal adversarial attacks, ensuring reduced latency and the generation of high-fidelity adversarial examples with superior success rates. Furthermore, we demonstrate that our framework achieves outstanding transferability and robustness against purification defenses, outperforming existing gradient-based attack models in both effectiveness and efficiency.
\end{abstract}

\begin{figure}[tb]
    \centering
    \includegraphics[width=1.1\columnwidth]{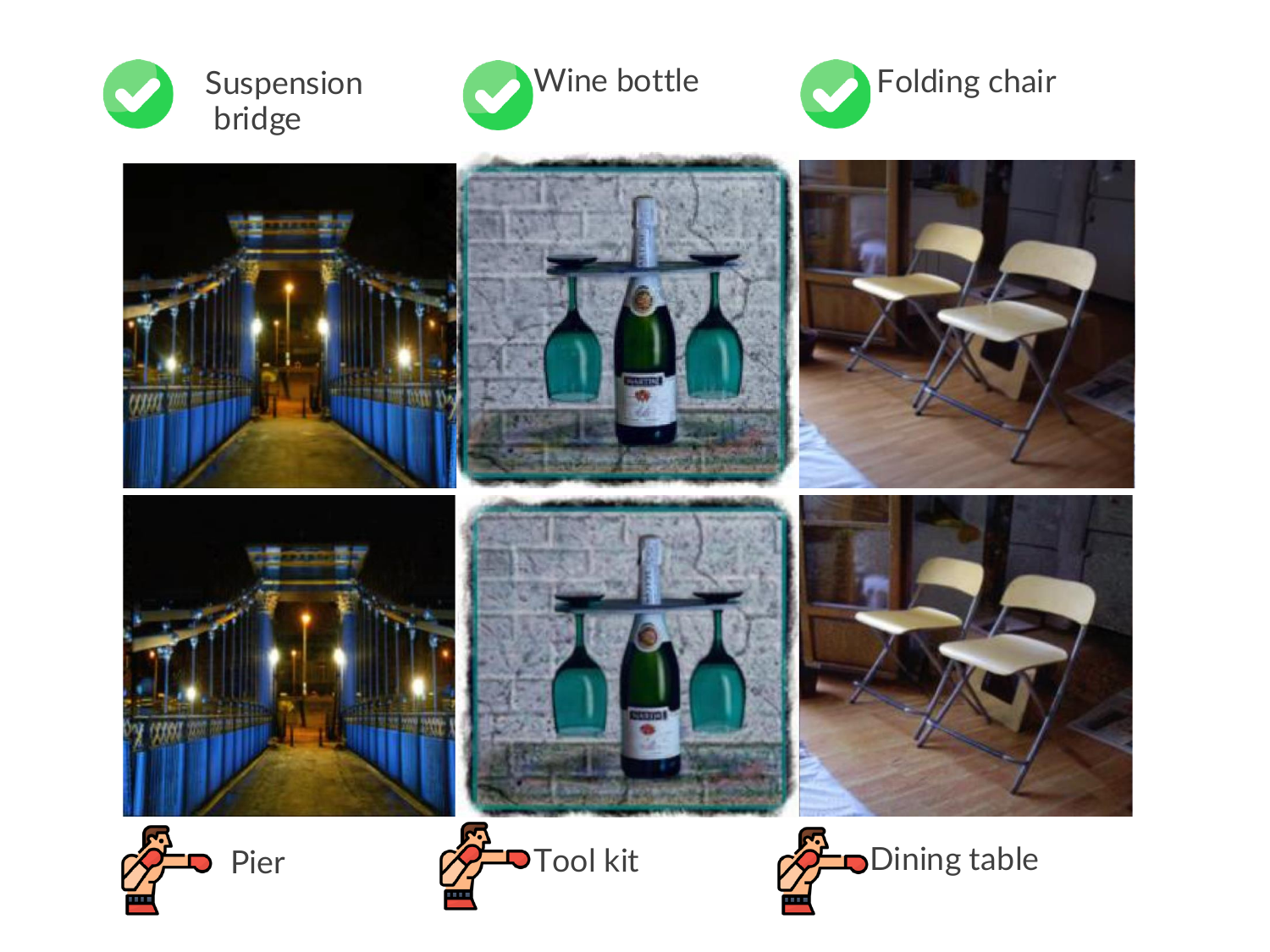}
    \caption{\textbf{Examples of MMAD-Purify.} The input image $\mathbf{x}$ is attacked by MMAD-Purify to create $\mathbf{x}_{\text{adv}}$, which is then purified to obtain $\mathbf{x}_{\text{adv}}^p$. The final sample demonstrates high image quality and robustness against defenses.}
    \label{fig:demo}
\end{figure}

\begin{figure*}
    \centering
    \includegraphics[width=\linewidth]{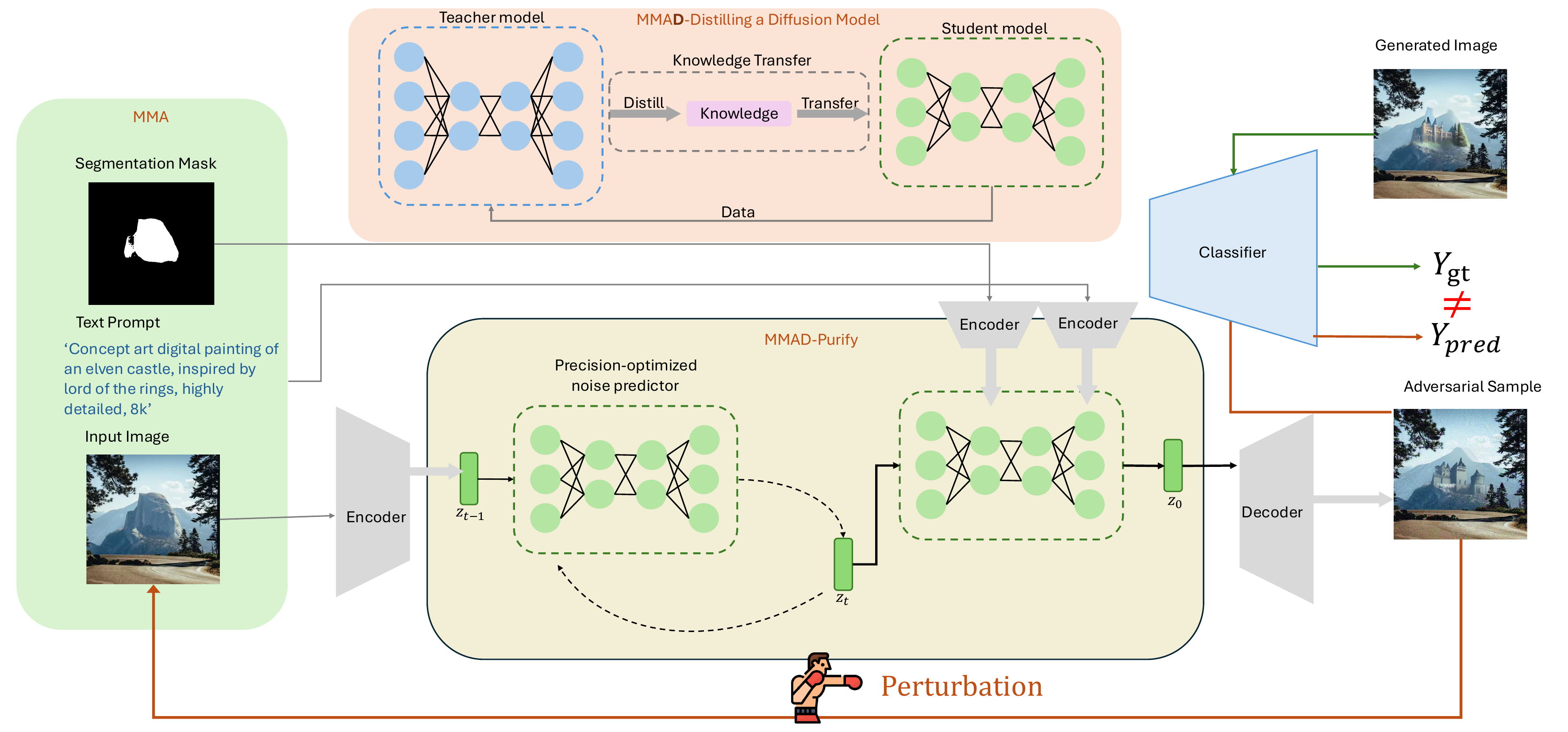}
    \caption{\textbf{Overview of MMAD-Purify}. In the MMAD-Purify framework, the input image is first processed through an encoder. It then enters the first distillated DM pipeline, where a precision-optimized noise predictor is applied. The resulting latents of the input image, combined with latents of other modalities, form a multi-modal representation. This multi-modal representation is then passed through a target classifier to generate perturbations, which are added back to the input image. This process is iteratively repeated, ultimately generating the adversarial example, $\mathbf{x}_{adv}$. After purification, the final $\mathbf{x}_{adv}^p$ has a different label from the generated image, demonstrating successful adversarial attack.}
    \label{fig:teaser}
\end{figure*}

\section{Introduction}
Artificial neural networks have demonstrated remarkable capabilities in automatically identifying relevant features in data and leveraging that learning to excel at a wide range of tasks. However, researchers have identified a significant vulnerability: their susceptibility to adversarial examples. These carefully crafted inputs, subtly modified in ways imperceptible to humans, can cause disruptive outputs from neural networks. This vulnerability raises serious security and reliability concerns, particularly affecting public trust in AI-enhanced critical systems such as autonomous vehicles and medical diagnostics~\cite{lau2023adversarial}.

In response to the adversarial vulnerabilities of neural networks, researchers have pursued two main directions. One focuses on developing robust defense mechanisms aimed at mitigating these vulnerabilities, ensuring that neural networks can perform reliably even under adversarial conditions~\cite{liu2022segment,dong2023enemy}. The other direction concentrates on crafting more sophisticated adversarial attacks, designed to identify and exploit weaknesses in existing defenses~\cite{sam-indentify-adv,sam-tifs-adv-training}. This ongoing "arms race" between attack and defense strategies has driven significant advancements in the field, contributing to the development of more secure and robust machine learning systems.

Despite these efforts, current research predominantly focuses on single-modality attacks, neglecting the diverse range of modalities such as text, 3D depth maps and optical flow. This narrow focus has led to a lack of robustness in models that incorporate these alternative modalities. To address this limitation and enhance overall model robustness, we propose a multi-modality adversarial attack framework. Our approach enables attacks across different modalities, providing a more comprehensive evaluation of model vulnerabilities.

Generative models have shown immense potential across various domains, with the emergence of diffusion models offering particular promise~\cite{ho2020denoising}~\cite{song2020score}. Compared to Generative Adversarial Networks (GANs), diffusion models demonstrate easier convergence and increased stability, leading to the development of numerous attack methods leveraging these generative capabilities. Given the outstanding performance of diffusion models, our multi-modality attack combines diffusion model techniques with traditional optimization-based gradient adversarial attack methods to generate adversarial samples. In implementing our multi-modality attack, we identified several challenges inherent to diffusion models in the context of adversarial sample generation. These include computational speed issues and the editability-reconstruction trade-off that emerges in multi-modal scenarios. To address these limitations, we employ distilled diffusion models and precision-optimized noise predictors. These techniques not only enhance the attack success rate but also improve the quality of adversarial samples. Furthermore, the increased transferability of the generated adversarial samples underscores the effectiveness and generalizability of our approach.

Our contributions can be summarized as follows:
\begin{itemize}
    \item To the best of our knowledge, we are the first to propose a multi-modality attack framework utilizing diffusion models, enabling enhanced robustness evaluation for a diverse range of models.
    \item We introduce an innovative approach that integrates a precision-optimized noise predictor scheme within the adversarial attack process, enhancing the purification of adversarial examples and yielding perturbations with high transferability.
    \item Our method substantially reduces inference time compared to conventional adversarial sample generation techniques that rely on diffusion models, making it applicable for real-world scenarios.
\end{itemize}

\section{Related Works}
\subsection{Adversarial Attack}
The discussion on the robustness and security of neural networks began with the seminal work by~\citeauthor{szegedy2013intriguing}~\cite{szegedy2013intriguing}. In their research, they discovered that neural networks are susceptible to small perturbations, which can easily fool the models and lead to misclassifications. This groundbreaking finding exposed a significant vulnerability in neural networks and sparked a new line of research focused on understanding and mitigating these weaknesses.
Following this initial discovery,~\citeauthor{goodfellow2014explaining} introduced the concept of adversarial examples and proposed the Fast Gradient Sign Method (FGSM) as a simple yet effective way to generate such examples~\cite{goodfellow2014explaining}.
Building upon the work of~\citeauthor{goodfellow2014explaining}, many new adversarial attack methods have also been proposed. The Projected Gradient Descent (PGD) attack is worth noting as an powerful attack method, which can generate deceptive adversarial examples by iteratively maximizing the model's loss function while constraining the perturbations within an $\epsilon$-ball~\cite{madry2017towards}.
\citeauthor{dong2018boosting} integrated momentum into the attack process, allowing for more stable and effective updates during the attack~\cite{dong2018boosting}.
\citeauthor{wang2021enhancing} introduced the concept of variance tuning, which involves adjusting the perturbation magnitude of adversarial examples, enhancing a higher transferability\cite{wang2021enhancing}.

The field of adversarial attacks has witnessed significant progress and has not only focused on image classification, but has also expanded beyond it~\cite{liu2023instruct2attack,liu2023diffprotect,guo2024white}.
These advancements have shed light on the vulnerabilities and challenges faced by deep learning models across different tasks and modalities, emphasizing the need for robust and comprehensive defense strategies.

\subsubsection{Latent Space Attack}
Instead of adding perturbations in the image space, latent space attacks perturb the latent representation of the input image, which is usually in the latent space of generative models.
\citeauthor{jalal2017robust} demonstrate the use of GANs and VAEs to create a new type of attack called the overpowered attack~\cite{jalal2017robust}. This attack searches over pairs of images in the spanner range, which improves upon the DefenseGAN.
\citeauthor{lau2023interpolated} propose a novel threat model called Interpolated Joint Space Adversarial Training (JSTM), which addresses perturbations in both image and latent spaces. Furthermore, to ensure JSA samples generalize well to real datasets, they exploit the invertibility of the Flow-based model, thus maintaining the exact manifold assumption~\cite{lau2023interpolated}.
Diff-Protect was proposed as a novel diffusion-based method for facial privacy protection~\cite{liu2023diffprotect}.
It generates meaningful perturbations in the latent space of the input images and produces adversarial images of high visual quality.

\subsection{Diffusion Model}
Diffusion models (DM) have excelled in image generation by learning to denoise corrupted data~\cite{ho2020denoising,song2020score,song2019generative}, offering improved stability and quality compared to VAEs and GANs~\cite{kingma2013auto,goodfellow2020generative}. However, DMs face challenges with slow inference. To address this, researchers have explored both training-free and training-based methods to accelerate the generation process. Training-free approaches include formulating denoising as an ODE and using adaptive step size solvers~\cite{song2019generative,jolicoeur2021gotta}, while training-based methods involve optimizing schedules, truncating the process, direct mapping, and knowledge distillation~\cite{watson2021learning,lyu2022accelerating,meng2023distillation}. Notably, Consistency Models have shown promise in enabling fast one-step generation while maintaining quality~\cite{song2023consistency,lcm}, representing a significant advancement in DM efficiency.

\section{Preliminaries}
Firstly, we will introduce the background of the continuous-time diffusion model. Let $\{\mathbf{X}(t)\}_{t \in [0, T]}$ be a diffusion process defined on the continuous time interval $[0, T]$, where $\mathbf{X}: [0, T] \times \Omega \rightarrow \mathbb{R}^d$ is a stochastic process. The forward process transitions the original data distribution $p_{\text{data}}(\mathbf{x})$ to a marginal distribution $p_t(\mathbf{x}_t)$ via the transition kernel $p_{0t}(\mathbf{x}_t | \mathbf{x}_0) = \mathcal{N}(\mathbf{x}_t | \alpha(t) \mathbf{x}_0, \sigma^2(t) \mathbf{I})$, where $\alpha, \sigma: [0,T] \rightarrow \mathbb{R}^+$ specify the noise schedule.
The dynamics of the forward process are governed by the following Itô stochastic differential equation (SDE):
\begin{equation}
\mathrm{d}\mathbf{X}(t) = \boldsymbol{\mu}(\mathbf{X}(t), t)\,\mathrm{d}t + g(t)\,\mathrm{d}\mathbf{W}(t), \quad t \in [0, T],
\end{equation}
where $\mathbf{W}(t)$ is a standard $d$-dimensional Wiener process, $\boldsymbol{\mu}: \mathbb{R}^d \times [0, T] \rightarrow \mathbb{R}^d$ is the drift coefficient, and $g: [0, T] \rightarrow \mathbb{R}^+$ is the diffusion coefficient.

To generate samples from the data distribution, we solve the reverse-time SDE:
\begin{align}
\mathrm{d} \mathbf{X}(t) = [\boldsymbol{\mu}(\mathbf{X}(t), &t)- \notag \\  g(t)^2& \nabla_{\mathbf{x}} \log p_t(\mathbf{X}(t))] \,\mathrm{d} t + 
g(t) \,\mathrm{d} \overline{\mathbf{W}}(t),
\end{align}
where $\overline{\mathbf{W}}(t)$ is a standard Wiener process in reverse time, starting from $\mathbf{X}(T) \sim p_T$.

The marginal distribution $p_t(\mathbf{x})$ satisfies the following ordinary differential equation, known as the Probability Flow ODE (PF-ODE)\cite{song2020score}:
\begin{equation}
\frac{\mathrm{d} \mathbf{x}_t}{\mathrm{d} t} = \boldsymbol{\mu}(\mathbf{x}_t, t) - \frac{1}{2} g^2(t) \nabla_{\mathbf{x}} \log p_t(\mathbf{x}_t), \quad \mathbf{x}_T \sim p_T(\mathbf{x}_T).
\end{equation}
In the diffusion model, we train a noise prediction model $\boldsymbol{\epsilon}_\theta$ to approximate the score function.

Classifier-free guidance (CFG)~\cite{ho2022classifier} is introduced to guide the sampling process towards a conditioning signal $\mathbf{c} $:
\begin{equation}
\hat{\boldsymbol{\epsilon}}_\theta(\mathbf{x}_t, \omega, \mathbf{c}, t) := (1+\omega) \boldsymbol{\epsilon}_\theta(\mathbf{x}_t, \mathbf{c}, t) - \omega \boldsymbol{\epsilon}_\theta(\mathbf{x}_t, \varnothing, t),
\end{equation}
where $\omega > 0$ is the guidance strength.

Approximating the score function by the noise prediction model, we obtain the following augmented PF-ODE~\cite{luo2023lcm} for sampling:
\begin{equation}
\begin{aligned}
    \frac{\mathrm{d} \mathbf{x}_t}{\mathrm{d} t}  = \boldsymbol{\mu}&(\mathbf{x}_t, t) + \frac{g^2(t)}{2\sigma(t)} \hat{\boldsymbol{\epsilon}}_\theta(\mathbf{x}_t, \omega, \mathbf{c}, t),\\
    &\textnormal{where}\,\, \mathbf{x}_T \sim \mathcal{N}(\mathbf{0}, \sigma^2(T) \mathbf{I}).
\end{aligned}
\end{equation}

\section{Method}
\subsection{Multi-Modalities Adversarial Attack(MMA)}
\label{subsec:sdxlpgd}

Let $\mathcal{X} \subseteq \mathbb{R}^D$ be the input space and $\mathcal{Y} = \{1, \ldots, K\}$ be the label space, where $K$ is the number of classes. Consider an image sample $\mathbf{x} \in \mathcal{X}$, a condition signal $\mathbf{c} \in \mathcal{C}$, and a true label $y \in \mathcal{Y}$. Let $h_{\psi}: \mathcal{X} \rightarrow \mathbb{R}^K$ be a target classifier with parameters $\psi$. The objective of our method is to craft an adversarial example $\mathbf{x}_{\text{adv}}$ that fools the target classifier $h_{\psi}$.  Let $(\mathcal{E}, \mathcal{D})$ be an autoencoder, where $\mathcal{E}: \mathcal{X} \rightarrow \mathcal{Z}$ and $\mathcal{D}: \mathcal{Z} \rightarrow \mathcal{X}$ are the encoder and decoder functions, respectively. Here, $\mathcal{X} \subseteq \mathbb{R}^D$ represents the input space, and $\mathcal{Z} \subseteq \mathbb{R}^d$ represents the latent space, with $d \ll D$. For any $\mathbf{x} \in \mathcal{X}$, we define the latent representation as $\mathbf{z} = \mathcal{E}(\mathbf{x})$ and the reconstructed output as $\tilde{\mathbf{x}} = \mathcal{D}(\mathbf{z})$.

For any $\mathbf{z} \in \mathcal{Z}$ and $n \in \{1, \ldots, N\}$, the noised latent variable $\mathbf{z}_{t_n}$ is defined by the following Gaussian distribution:
\begin{equation}\label{forward}
\mathbf{z}_{t_n} \sim \mathcal{N}\left(\alpha(t_n) \mathbf{z}, \sigma^2(t_n) \mathbf{I}\right),
\end{equation}
where $\alpha: [0, T] \rightarrow \mathbb{R}^+$ and $\sigma: [0, T] \rightarrow \mathbb{R}^+$ are the noise schedule functions.

Consider a partition of the time interval $[0, T]$ into $N-1$ subintervals: $0=t_1<t_2<\cdots<t_{N-1}<t_N=T$. Let $\Phi: \mathcal{Z} \times[0, T]^2 \times \mathcal{C} \times \mathcal{Z} \times \Theta \rightarrow \mathcal{Z}$ be a one-step numerical solver for the augmented Probability Flow ODE (PF-ODE), where $\Theta$ denotes the parameter space. For any $n \in\{1, \ldots, N-1\}$, we define the one-step estimate of the latent variable $\mathbf{z}_{t_{n-1}}$ from $\mathbf{z}_{t_n}$ using the numerical solver as follows:
\begin{equation}\label{each_step_solver}
\mathbf{z}_{t_{n-1}}=\Phi\left(\mathbf{z}_{t_{n}}, t_n, t_{n-1}, \mathbf{c}, \boldsymbol{\epsilon_{\theta}} ; \boldsymbol{\phi}\right),
\end{equation}
where $\boldsymbol{\phi} \in \Theta$ represents the parameters of the numerical solver, $\epsilon_\theta$ is the noise predictor.

 Now we obtain the latent vector through recursive application of the one-step estimates at $t= 0$ from $t = \tau$:
\begin{equation}
\tilde{\mathbf{z}} := (\tilde{\mathbf{z}}_{t_1}^\Phi \circ \tilde{\mathbf{z}}_{t_2}^\Phi \circ \cdots \circ \tilde{\mathbf{z}}_{t_{n-1}}^\Phi)(\mathbf{z}_{\tau}).
\label{recursive}
\end{equation}
Finally, the reconstructed image estimate is given by:
\begin{equation}
\tilde{\mathbf{x}} = \mathcal{D}(\tilde{\mathbf{z}}).
\end{equation}

We define a loss function $\ell: \mathbb{R}^K \times \mathcal{Y} \to \mathbb{R}_{\geq 0}$ that quantifies the discrepancy between the model's prediction $h_{\psi}(\mathbf{x})$ and the true label $y \in \mathcal{Y}$.

The objective of an $\ell_{\infty}$-norm constrained adversarial attack is to maximize the loss $\ell(h_{\psi}(\tilde{\mathbf{x}}(\mathbf{x} + \boldsymbol{\delta})), y)$ with respect to the perturbation $\boldsymbol{\delta}$, subject to $\|\boldsymbol{\delta}\|_{\infty} \leq \epsilon$ for some $\epsilon > 0$:

\begin{equation}
\max_{\boldsymbol{\delta} \in \mathbb{R}^D} \ell(h_{\psi}(\tilde{\mathbf{x}}(\mathbf{x} + \boldsymbol{\delta})), y) \quad \text{subject to} \quad \|\boldsymbol{\delta}\|_{\infty} \leq \epsilon
\end{equation}

To enforce the $\ell_{\infty}$ constraint, we define an $\epsilon$-ball projection operator $\Pi_{\epsilon}$ that clips the perturbation element-wise:
\begin{equation}
\Pi_{\epsilon}(\boldsymbol{\delta}) = \min(\max(\boldsymbol{\delta}, -\epsilon \mathbf{1}), \epsilon \mathbf{1}),
\end{equation}
where $\mathbf{1} \in \mathbb{R}^D$ is the all-ones vector and the $\min$ and $\max$ operations are applied element-wise.

Let $\boldsymbol{\delta}^{(i)}$ denote the perturbation at iteration $i$. At each step, we take a gradient ascent step to maximize the loss, followed by projecting the perturbation onto the $\epsilon$-ball:
\begin{equation}
\begin{aligned}
\hat{\boldsymbol{\delta}}^{(i)} &= \boldsymbol{\delta}^{(i-1)} + \eta \nabla_{\boldsymbol{\delta}} \ell(h_{\psi}(\tilde{\mathbf{x}}(\mathbf{x} + \boldsymbol{\delta}^{(i-1)})), y), \\
\boldsymbol{\delta}^{(i)} &= \Pi_{\epsilon}(\hat{\boldsymbol{\delta}}^{(i)}),
\end{aligned}
\end{equation}
where $\eta > 0$ is the step size and $\tilde{\mathbf{x}}^{(i-1)}$ is the reconstructed image estimate at iteration $i-1$. After $I$ iterations, the final adversarial perturbation is given by $\boldsymbol{\delta}^{(I)}$, and the corresponding adversarial example is:
\begin{equation}
\mathbf{x}_{\text{adv}} := \mathbf{x} + \boldsymbol{\delta}^{(I)}.
\end{equation}

\subsection{Speeding Up Multi-modality Adversarial Attacks with a Distilled Diffusion Model (MMAD)}

However, we observe that the iterative nature of both the diffusion model and the PGD attack presents a critical limitation: the process is considerably slow and requires substantial computational resources. To address this inefficiency, we employ a multi-sampling strategy within our framework by integrating a distilled diffusion model. This methodology leverages a pre-trained diffusion model as a teacher to guide the distillation of a student model. By doing so, we optimize the reconstruction process in our attack framework, condensing it to just 1 to 4 steps, thereby markedly improving the computational efficiency of our attack. Further details can be found in the Appendix. 

We replace the recursive application of one-step estimates with a single-step prediction using a distilled diffusion model. Specifically, we define $\boldsymbol{f}_{\boldsymbol{\theta}}: \mathcal{Z} \times \mathbb{R}^+\times \mathcal{C} \times [0, T] \rightarrow \mathcal{Z}$ as our distilled model function, which efficiently predicts the solution of the Probability Flow ODE at $t = 0$. We replace Equation \eqref{recursive} with the following:

\begin{equation}
\tilde{\mathbf{z}} = \boldsymbol{f}_\theta\left(\mathbf{z}_{t_n}, \omega, \mathbf{c}, t_n\right)
\label{distilled_model}
\end{equation}

where $\mathbf{z}_{t_n}$ is the noisy latent variable at time $t_n$, $\omega$ is the guidance strength, $\mathbf{c}$ is the condition, and $t_n$ is the current time step.

This replacement reduces the reconstruction process from multiple recursive steps to a single prediction operation, greatly improving computational efficiency. By utilizing this distilled model, we optimize the reconstruction process to require only 1 to 4 steps, significantly enhancing the computational efficiency of our attack method.

\begin{table}[ht]
\centering
\resizebox{0.5\columnwidth}{!}{%
\begin{tabular}{cccc}
\hline
        & Diff-PGD &  MMAD\\ \hline
runtime & 6.3s     &  \textbf{2.0s}           \\ \hline
\end{tabular}%
}
\caption{Compare of the runtime of Diff-PGD, and the proposed MMAD. The experiment is taken on a NVIDIA RTX3060, parameters follow or are equivalent to the setting of Diff-PGD.}
\label{tab:run-time-sdxl}
\end{table}

\subsection{Strengthening MMAD with Precision-Optimized Noise Predictor(MMAD-Purify)}

Although Multi-Modal Adversarial Diffusion (MMAD) improves the speed of the attack, we observe that it has lower attack accuracy. We identify two main reasons for this limitation:

1) Violation of the linearity assumption: Diffusion models typically rely on a small step size \(\Delta t\) to maintain a nearly linear trajectory throughout the diffusion process. However, MMAD uses a distilled model, which compresses the number of steps required for image reconstruction down to just four. This significant reduction in steps causes the trajectory to deviate considerably from linearity, leading to less accurate reconstructions and, consequently, a reduction in attack efficacy.

2) Multi-modality trade-off: While our method integrates multi-modality to enhance attack versatility, it introduces a limitation due to the inherent trade-off between reconstruction accuracy and editability in diffusion models~\cite{tov2021designing,garibi2024renoise}. When attempting to balance multiple modalities, the model may prioritize editability over precise reconstruction, further compromising the accuracy of the attack. As a result, the model's ability to generate consistent and accurate adversarial examples is diminished.

To address these limitations and mitigate the trade-off between reconstruction accuracy and editability in diffusion models, we employ regularization techniques inspired by ReNoise~\cite{garibi2024renoise} and Zero-Shot~\cite{parmar2023zero}. These techniques involve two key components: the pairwise correlation loss \(\mathcal{L}_{\text{pair}}\) and the patch-wise KL-divergence loss \(\mathcal{L}_{\text{patch-KL}}\). The pairwise correlation loss reduces correlations between predicted noise values across pixel pairs, encouraging the generation of noise patterns that approximate the characteristics of uncorrelated Gaussian white noise. Meanwhile, the patch-wise KL-divergence loss aligns the predicted noise during reverse diffusion with the noise from the forward process, ensuring consistency in the diffusion process. These two regularization losses are combined into a single editability loss, \(\mathcal{L}_{\text{edit}}\), which balances the reconstruction accuracy and editability, thereby improving the overall effectiveness of the attack.

As discussed in section of MMA~\ref{subsec:sdxlpgd}, we now modify the step in Equation~\ref{forward}. Let \(\epsilon_\theta: \mathcal{Z} \times [0, T] \times \mathcal{C} \times \mathbb{R}^+ \rightarrow \mathcal{Z}\) be the noise prediction function parameterized by \(\theta\). Define \(\Psi: \mathcal{Z} \times [0, T]^2 \times \mathcal{C} \times \mathcal{Z} \times \Theta \rightarrow \mathcal{Z}\) as the step function, which iteratively refines the estimate of the latent variable at each step, guided by a series of weights \(\left\{\alpha_\ell\right\}_{\ell=1}^{L}\). Consider the sequences \(\left\{\mathbf{z}_{t_{n}}^{(\ell)}\right\}_{\ell=0}^{L}\) and \(\left\{\boldsymbol{\epsilon}_\theta^{(\ell)}\right\}_{\ell=0}^{L}\) defined as follows:

For each \(\ell \in \{0, 1, \ldots, L\}\), the noise prediction \(\boldsymbol{\epsilon}_\theta^{(\ell)}\) is given by:
\begin{equation}
\boldsymbol{\epsilon}_\theta^{(\ell)} = \epsilon_\theta\left(\mathbf{z}_{t_{n}}^{(\ell)}, t_n, \mathbf{c}, \mathbf{\omega}\right),
\end{equation}
before updating the sequence, we regularize the noise prediction \(\boldsymbol{\epsilon}_\theta^{(\ell)}\) using the editability loss \(\mathcal{L}_{\text{edit}}\). The regularized noise prediction \(\boldsymbol{\epsilon}_{\theta,\text{reg}}^{(\ell)} \in \mathbb{R}^n\) at iteration \(\ell\) is then defined as:
\begin{equation}
\boldsymbol{\epsilon}_{\theta,\text{reg}}^{(\ell)} = \boldsymbol{\epsilon}_\theta^{(\ell)} + \begin{cases}
-\nabla_{\boldsymbol{\epsilon}} \mathcal{L}_{\text{edit}}(\boldsymbol{\epsilon}_\theta^{(\ell)}), & \text{if } \alpha_\ell > 0\\
0, & \text{otherwise}
\end{cases}
,
\end{equation}
where \(\mathcal{L}_{\text{edit}} = \lambda_{\text{patch-KL}} \mathcal{L}_{\text{patch-KL}} + \lambda_{\text{pair}} \mathcal{L}_{\text{pair}}\).

Define the average noise prediction \(\overline{\boldsymbol{\epsilon}}^{(\ell)}\) as:
\begin{equation}
\overline{\boldsymbol{\epsilon}}^{(\ell)} = \frac{\ell \cdot \overline{\boldsymbol{\epsilon}}^{(\ell-1)} + \boldsymbol{\epsilon}_{\theta, \text{reg}}^{(\ell)}}{\ell + 1},
\end{equation}
using this regularized average noise prediction, define the sequence of estimates \(\left\{\mathbf{z}_{t_{n}}^{(\ell+1)}\right\}_{\ell=0}^{L}\) through the inverse step function \(\Psi\) as:
\begin{equation}
\mathbf{z}_{t_{n}}^{(\ell+1)} = \Psi\left(\mathbf{z}_{t_{n-1}}^{(\ell)}, t_n, t_{n-1}, \mathbf{c}, \boldsymbol{\epsilon}_{\theta, \text{reg}}^{(\ell)}; \boldsymbol{\psi}\right).
\end{equation}

The final estimation of the latent variable at time \(t_{n}\) is then given by:
\begin{equation}
\hat{\mathbf{z}}_{t_{n}} = \mathbf{z}_{t_{n}}^{(L)}.
\end{equation}

Now, we continue Equation~\ref{distilled_model}, and following steps in MMA section get $\mathbf{x}_{\text{adv}}$. To enhance the robustness of our adversarial attack, we finally use $\mathbf{x}_{\text{adv}}$ as the input for the distilled diffusion model regularized by precision optimized noise predictor. The final output, $\mathbf{x}_{\text{adv}}^p$, is designed to exhibit increased robustness.Figure~\ref{fig:demo} showcases exemplary results generated by our method, while Figure~\ref{fig:teaser} illustrates the comprehensive overview of MMAD-Purify framework.

\begin{table*}[tb]
\centering
\resizebox{\textwidth}{!}{%
\begin{tabular}{ccccccccccccc}
\hline
                                               & \multicolumn{2}{c}{}                                                          & \multicolumn{6}{c}{Neural Networks}                                                                                                                                                                                                                                                                                   & \multicolumn{4}{c}{IQAs}                                                                                       \\
\multirow{-2}{*}{White-box Models}             & \multicolumn{2}{c}{\multirow{-2}{*}{Attack Methods}}                          & VGG19                                 & RN50                                                          & WR101                                                         & DN121                                                         & NNv2                                  & SNv2                                  & PSNR $\uparrow$              & SSIM $\uparrow$             & FID $\downarrow$             & LPIPS $\downarrow$ \\ \hline
\multicolumn{1}{c|}{}                          & \multicolumn{1}{c|}{}                         & \multicolumn{1}{c|}{PGD}      & \multicolumn{1}{c|}{50.39\%}          & \multicolumn{1}{c|}{\cellcolor[HTML]{C0C0C0}35.37\%}          & \multicolumn{1}{c|}{30.82\%}                                  & \multicolumn{1}{c|}{39.71\%}                                  & \multicolumn{1}{c|}{47.33\%}          & \multicolumn{1}{c|}{51.43\%}          & \multicolumn{1}{c|}{28.5391} & \multicolumn{1}{c|}{0.8370} & \multicolumn{1}{c|}{14.9556} & 0.1738             \\
\multicolumn{1}{c|}{}                          & \multicolumn{1}{c|}{}                         & \multicolumn{1}{c|}{Diff-PGD} & \multicolumn{1}{c|}{55.47\%}          & \multicolumn{1}{c|}{\cellcolor[HTML]{C0C0C0}\hl{78.98\%}} & \multicolumn{1}{c|}{39.43\%}                                  & \multicolumn{1}{c|}{44.88\%}                                  & \multicolumn{1}{c|}{56.16\%}          & \multicolumn{1}{c|}{55.86\%}          & \multicolumn{1}{c|}{28.3815} & \multicolumn{1}{c|}{0.8293} & \multicolumn{1}{c|}{17.1181} & 0.1842             \\
\multicolumn{1}{c|}{}                          & \multicolumn{1}{c|}{\multirow{-3}{*}{$p_1$}}  & \multicolumn{1}{c|}{Ours}     & \multicolumn{1}{c|}{\hl{58.76\%}} & \multicolumn{1}{c|}{\cellcolor[HTML]{C0C0C0}69.50\%}          & \multicolumn{1}{c|}{\hl{46.67\%}}                         & \multicolumn{1}{c|}{\hl{49.77\%}}                         & \multicolumn{1}{c|}{\hl{58.80\%}} & \multicolumn{1}{c|}{\hl{60.45\%}} & \multicolumn{1}{c|}{28.2141} & \multicolumn{1}{c|}{0.8194} & \multicolumn{1}{c|}{19.1739} & 0.1964             \\ \cline{2-13} 
\multicolumn{1}{c|}{}                          & \multicolumn{1}{c|}{}                         & \multicolumn{1}{c|}{PGD}      & \multicolumn{1}{c|}{46.85\%}          & \multicolumn{1}{c|}{\cellcolor[HTML]{C0C0C0}23.85\%}          & \multicolumn{1}{c|}{26.03\%}                                  & \multicolumn{1}{c|}{35.06\%}                                  & \multicolumn{1}{c|}{45.42\%}          & \multicolumn{1}{c|}{49.18\%}          & \multicolumn{1}{c|}{27.2513} & \multicolumn{1}{c|}{0.8041} & \multicolumn{1}{c|}{18.9602} & 0.1990             \\
\multicolumn{1}{c|}{}                          & \multicolumn{1}{c|}{}                         & \multicolumn{1}{c|}{Diff-PGD} & \multicolumn{1}{c|}{49.11\%}          & \multicolumn{1}{c|}{\cellcolor[HTML]{C0C0C0}\hl{46.05\%}} & \multicolumn{1}{c|}{27.57\%}                                  & \multicolumn{1}{c|}{37.34\%}                                  & \multicolumn{1}{c|}{44.28\%}          & \multicolumn{1}{c|}{50.00\%}          & \multicolumn{1}{c|}{27.1843} & \multicolumn{1}{c|}{0.8013} & \multicolumn{1}{c|}{19.4638} & 0.2013             \\
\multicolumn{1}{c|}{}                          & \multicolumn{1}{c|}{\multirow{-3}{*}{$p_1+$}} & \multicolumn{1}{c|}{Ours}     & \multicolumn{1}{c|}{\hl{49.68\%}} & \multicolumn{1}{c|}{\cellcolor[HTML]{C0C0C0}44.52\%}          & \multicolumn{1}{c|}{\hl{36.04\%}}                         & \multicolumn{1}{c|}{\hl{41.27\%}}                         & \multicolumn{1}{c|}{\hl{50.08\%}} & \multicolumn{1}{c|}{\hl{53.31\%}} & \multicolumn{1}{c|}{27.1135} & \multicolumn{1}{c|}{0.7968} & \multicolumn{1}{c|}{20.9897} & 0.2049             \\ \cline{2-13} 
\multicolumn{1}{c|}{}                          & \multicolumn{1}{c|}{}                         & \multicolumn{1}{c|}{PGD}      & \multicolumn{1}{c|}{64.57\%}          & \multicolumn{1}{c|}{\cellcolor[HTML]{C0C0C0}43.49\%}          & \multicolumn{1}{c|}{48.17\%}                                  & \multicolumn{1}{c|}{49.37\%}                                  & \multicolumn{1}{c|}{65.65\%}          & \multicolumn{1}{c|}{64.75\%}          & \multicolumn{1}{c|}{26.7452} & \multicolumn{1}{c|}{0.8125} & \multicolumn{1}{c|}{29.1990} & 0.2280             \\
\multicolumn{1}{c|}{}                          & \multicolumn{1}{c|}{}                         & \multicolumn{1}{c|}{Diff-PGD} & \multicolumn{1}{c|}{66.42\%}          & \multicolumn{1}{c|}{\cellcolor[HTML]{C0C0C0}68.27\%}          & \multicolumn{1}{c|}{47.14\%}                                  & \multicolumn{1}{c|}{55.12\%}                                  & \multicolumn{1}{c|}{67.16\%}          & \multicolumn{1}{c|}{67.43\%}          & \multicolumn{1}{c|}{26.5719} & \multicolumn{1}{c|}{0.8036} & \multicolumn{1}{c|}{31.5850} & 0.2415             \\
\multicolumn{1}{c|}{}                          & \multicolumn{1}{c|}{\multirow{-3}{*}{$p_2$}}  & \multicolumn{1}{c|}{Ours}     & \multicolumn{1}{c|}{\hl{80.25\%}} & \multicolumn{1}{c|}{\cellcolor[HTML]{C0C0C0}\hl{95.50\%}} & \multicolumn{1}{c|}{\hl{75.68\%}}                         & \multicolumn{1}{c|}{\hl{73.60\%}}                         & \multicolumn{1}{c|}{\hl{80.13\%}} & \multicolumn{1}{c|}{\hl{75.61\%}} & \multicolumn{1}{c|}{26.4921} & \multicolumn{1}{c|}{0.7970} & \multicolumn{1}{c|}{36.2468} & 0.2549             \\ \cline{2-13} 
\multicolumn{1}{c|}{}                          & \multicolumn{1}{c|}{}                         & \multicolumn{1}{c|}{PGD}      & \multicolumn{1}{c|}{65.35\%}          & \multicolumn{1}{c|}{\cellcolor[HTML]{C0C0C0}42.38\%}          & \multicolumn{1}{c|}{53.20\%}                                  & \multicolumn{1}{c|}{56.17\%}                                  & \multicolumn{1}{c|}{71.76\%}          & \multicolumn{1}{c|}{60.66\%}          & \multicolumn{1}{c|}{24.8956} & \multicolumn{1}{c|}{0.7744} & \multicolumn{1}{c|}{38.6395} & 0.2715             \\
\multicolumn{1}{c|}{}                          & \multicolumn{1}{c|}{}                         & \multicolumn{1}{c|}{Diff-PGD} & \multicolumn{1}{c|}{65.38\%}          & \multicolumn{1}{c|}{\cellcolor[HTML]{C0C0C0}56.06\%}          & \multicolumn{1}{c|}{51.86\%}                                  & \multicolumn{1}{c|}{55.63\%}                                  & \multicolumn{1}{c|}{69.65\%}          & \multicolumn{1}{c|}{63.36\%}          & \multicolumn{1}{c|}{24.7742} & \multicolumn{1}{c|}{0.7678} & \multicolumn{1}{c|}{41.6833} & 0.2811             \\
\multicolumn{1}{c|}{\multirow{-12}{*}{RN50}}   & \multicolumn{1}{c|}{\multirow{-3}{*}{$p_2+$}} & \multicolumn{1}{c|}{Ours}     & \multicolumn{1}{c|}{\hl{76.27\%}} & \multicolumn{1}{c|}{\cellcolor[HTML]{C0C0C0}\hl{88.95\%}} & \multicolumn{1}{c|}{\hl{72.97\%}}                         & \multicolumn{1}{c|}{\hl{70.71\%}}                         & \multicolumn{1}{c|}{\hl{79.81\%}} & \multicolumn{1}{c|}{\hl{73.17\%}} & \multicolumn{1}{c|}{24.7144} & \multicolumn{1}{c|}{0.7628} & \multicolumn{1}{c|}{44.6224} & 0.2906             \\ \hline
\multicolumn{1}{c|}{}                          & \multicolumn{1}{c|}{}                         & \multicolumn{1}{c|}{PGD}      & \multicolumn{1}{c|}{51.92\%}          & \multicolumn{1}{c|}{36.17\%}                                  & \multicolumn{1}{c|}{\cellcolor[HTML]{C0C0C0}34.67\%}          & \multicolumn{1}{c|}{41.14\%}                                  & \multicolumn{1}{c|}{50.31\%}          & \multicolumn{1}{c|}{57.39\%}          & \multicolumn{1}{c|}{28.5138} & \multicolumn{1}{c|}{0.8353} & \multicolumn{1}{c|}{14.8201} & 0.1754             \\
\multicolumn{1}{c|}{}                          & \multicolumn{1}{c|}{}                         & \multicolumn{1}{c|}{Diff-PGD} & \multicolumn{1}{c|}{55.83\%}          & \multicolumn{1}{c|}{44.81\%}                                  & \multicolumn{1}{c|}{\cellcolor[HTML]{C0C0C0}\hl{75.53\%}} & \multicolumn{1}{c|}{49.92\%}                                  & \multicolumn{1}{c|}{57.27\%}          & \multicolumn{1}{c|}{58.33\%}          & \multicolumn{1}{c|}{28.3655} & \multicolumn{1}{c|}{0.8280} & \multicolumn{1}{c|}{16.7484} & 0.1853             \\
\multicolumn{1}{c|}{}                          & \multicolumn{1}{c|}{\multirow{-3}{*}{$p_1$}}  & \multicolumn{1}{c|}{Ours}     & \multicolumn{1}{c|}{\hl{61.39\%}} & \multicolumn{1}{c|}{\hl{50.66\%}}                         & \multicolumn{1}{c|}{\cellcolor[HTML]{C0C0C0}62.39\%}          & \multicolumn{1}{c|}{\hl{51.24\%}}                         & \multicolumn{1}{c|}{\hl{59.39\%}} & \multicolumn{1}{c|}{\hl{58.96\%}} & \multicolumn{1}{c|}{28.1855} & \multicolumn{1}{c|}{0.8173} & \multicolumn{1}{c|}{19.0033} & 0.1983             \\ \cline{2-13} 
\multicolumn{1}{c|}{}                          & \multicolumn{1}{c|}{}                         & \multicolumn{1}{c|}{PGD}      & \multicolumn{1}{c|}{46.37\%}          & \multicolumn{1}{c|}{32.01\%}                                  & \multicolumn{1}{c|}{\cellcolor[HTML]{C0C0C0}20.04\%}          & \multicolumn{1}{c|}{38.90\%}                                  & \multicolumn{1}{c|}{44.44\%}          & \multicolumn{1}{c|}{\hl{56.09\%}} & \multicolumn{1}{c|}{27.2350} & \multicolumn{1}{c|}{0.8033} & \multicolumn{1}{c|}{18.6536} & 0.1993             \\
\multicolumn{1}{c|}{}                          & \multicolumn{1}{c|}{}                         & \multicolumn{1}{c|}{Diff-PGD} & \multicolumn{1}{c|}{47.48\%}          & \multicolumn{1}{c|}{36.31\%}                                  & \multicolumn{1}{c|}{\cellcolor[HTML]{C0C0C0}\hl{41.52\%}} & \multicolumn{1}{c|}{38.33\%}                                  & \multicolumn{1}{c|}{46.71\%}          & \multicolumn{1}{c|}{52.59\%}          & \multicolumn{1}{c|}{27.1764} & \multicolumn{1}{c|}{0.8007} & \multicolumn{1}{c|}{19.3679} & 0.2016             \\
\multicolumn{1}{c|}{}                          & \multicolumn{1}{c|}{\multirow{-3}{*}{$p_1+$}} & \multicolumn{1}{c|}{Ours}     & \multicolumn{1}{c|}{\hl{54.24\%}} & \multicolumn{1}{c|}{\hl{39.60\%}}                         & \multicolumn{1}{c|}{\cellcolor[HTML]{C0C0C0}39.18\%}          & \multicolumn{1}{c|}{\hl{39.93\%}}                         & \multicolumn{1}{c|}{\hl{51.26\%}} & \multicolumn{1}{c|}{53.73\%}          & \multicolumn{1}{c|}{27.0974} & \multicolumn{1}{c|}{0.7956} & \multicolumn{1}{c|}{20.6378} & 0.2065             \\ \cline{2-13} 
\multicolumn{1}{c|}{}                          & \multicolumn{1}{c|}{}                         & \multicolumn{1}{c|}{PGD}      & \multicolumn{1}{c|}{66.03\%}          & \multicolumn{1}{c|}{48.46\%}                                  & \multicolumn{1}{c|}{\cellcolor[HTML]{C0C0C0}39.08\%}          & \multicolumn{1}{c|}{52.34\%}                                  & \multicolumn{1}{c|}{65.20\%}          & \multicolumn{1}{c|}{64.35\%}          & \multicolumn{1}{c|}{26.7511} & \multicolumn{1}{c|}{0.8125} & \multicolumn{1}{c|}{28.7923} & 0.2289             \\
\multicolumn{1}{c|}{}                          & \multicolumn{1}{c|}{}                         & \multicolumn{1}{c|}{Diff-PGD} & \multicolumn{1}{c|}{68.70\%}          & \multicolumn{1}{c|}{54.18\%}                                  & \multicolumn{1}{c|}{\cellcolor[HTML]{C0C0C0}62.29\%}          & \multicolumn{1}{c|}{56.26\%}                                  & \multicolumn{1}{c|}{67.65\%}          & \multicolumn{1}{c|}{68.89\%}          & \multicolumn{1}{c|}{26.5408} & \multicolumn{1}{c|}{0.8027} & \multicolumn{1}{c|}{31.5989} & 0.2434             \\
\multicolumn{1}{c|}{}                          & \multicolumn{1}{c|}{\multirow{-3}{*}{$p_2$}}  & \multicolumn{1}{c|}{Ours}     & \multicolumn{1}{c|}{\hl{80.60\%}} & \multicolumn{1}{c|}{\hl{74.59\%}}                         & \multicolumn{1}{c|}{\cellcolor[HTML]{C0C0C0}\hl{95.80\%}} & \multicolumn{1}{c|}{\hl{71.91\%}}                         & \multicolumn{1}{c|}{\hl{82.67\%}} & \multicolumn{1}{c|}{\hl{78.36\%}} & \multicolumn{1}{c|}{26.4814} & \multicolumn{1}{c|}{0.7970} & \multicolumn{1}{c|}{35.1437} & 0.2528             \\ \cline{2-13} 
\multicolumn{1}{c|}{}                          & \multicolumn{1}{c|}{}                         & \multicolumn{1}{c|}{PGD}      & \multicolumn{1}{c|}{63.68\%}          & \multicolumn{1}{c|}{53.16\%}                                  & \multicolumn{1}{c|}{\cellcolor[HTML]{C0C0C0}39.18\%}          & \multicolumn{1}{c|}{54.18\%}                                  & \multicolumn{1}{c|}{70.44\%}          & \multicolumn{1}{c|}{62.39\%}          & \multicolumn{1}{c|}{24.9099} & \multicolumn{1}{c|}{0.7749} & \multicolumn{1}{c|}{38.1896} & 0.2714             \\
\multicolumn{1}{c|}{}                          & \multicolumn{1}{c|}{}                         & \multicolumn{1}{c|}{Diff-PGD} & \multicolumn{1}{c|}{69.22\%}          & \multicolumn{1}{c|}{56.05\%}                                  & \multicolumn{1}{c|}{\cellcolor[HTML]{C0C0C0}50.05\%}          & \multicolumn{1}{c|}{59.66\%}                                  & \multicolumn{1}{c|}{70.76\%}          & \multicolumn{1}{c|}{62.78\%}          & \multicolumn{1}{c|}{24.7410} & \multicolumn{1}{c|}{0.7667} & \multicolumn{1}{c|}{41.8474} & 0.2831             \\
\multicolumn{1}{c|}{\multirow{-12}{*}{WRN101}} & \multicolumn{1}{c|}{\multirow{-3}{*}{$p_2+$}} & \multicolumn{1}{c|}{Ours}     & \multicolumn{1}{c|}{\hl{75.33\%}} & \multicolumn{1}{c|}{\hl{72.28\%}}                         & \multicolumn{1}{c|}{\cellcolor[HTML]{C0C0C0}\hl{86.03\%}} & \multicolumn{1}{c|}{\hl{70.85\%}}                         & \multicolumn{1}{c|}{\hl{81.77\%}} & \multicolumn{1}{c|}{\hl{72.57\%}} & \multicolumn{1}{c|}{24.7115} & \multicolumn{1}{c|}{0.7633} & \multicolumn{1}{c|}{43.9790} & 0.2887             \\ \hline
\multicolumn{1}{c|}{}                          & \multicolumn{1}{c|}{}                         & \multicolumn{1}{c|}{PGD}      & \multicolumn{1}{c|}{46.92\%}          & \multicolumn{1}{c|}{35.62\%}                                  & \multicolumn{1}{c|}{32.59\%}                                  & \multicolumn{1}{c|}{\cellcolor[HTML]{C0C0C0}52.00\%}          & \multicolumn{1}{c|}{45.16\%}          & \multicolumn{1}{c|}{52.77\%}          & \multicolumn{1}{c|}{28.5101} & \multicolumn{1}{c|}{0.8358} & \multicolumn{1}{c|}{15.4706} & 0.1751             \\
\multicolumn{1}{c|}{}                          & \multicolumn{1}{c|}{}                         & \multicolumn{1}{c|}{Diff-PGD} & \multicolumn{1}{c|}{55.95\%}          & \multicolumn{1}{c|}{46.84\%}                                  & \multicolumn{1}{c|}{41.76\%}                                  & \multicolumn{1}{c|}{\cellcolor[HTML]{C0C0C0}\hl{89.80\%}} & \multicolumn{1}{c|}{55.65\%}          & \multicolumn{1}{c|}{57.47\%}          & \multicolumn{1}{c|}{28.3412} & \multicolumn{1}{c|}{0.8286} & \multicolumn{1}{c|}{18.5986} & 0.1860             \\
\multicolumn{1}{c|}{}                          & \multicolumn{1}{c|}{\multirow{-3}{*}{$p_1$}}  & \multicolumn{1}{c|}{Ours}     & \multicolumn{1}{c|}{\hl{60.64\%}} & \multicolumn{1}{c|}{\hl{52.52\%}}                         & \multicolumn{1}{c|}{\hl{48.85\%}}                         & \multicolumn{1}{c|}{\cellcolor[HTML]{C0C0C0}77.79\%}          & \multicolumn{1}{c|}{\hl{56.60\%}} & \multicolumn{1}{c|}{\hl{59.04\%}} & \multicolumn{1}{c|}{28.1799} & \multicolumn{1}{c|}{0.8189} & \multicolumn{1}{c|}{20.7998} & 0.1988             \\ \cline{2-13} 
\multicolumn{1}{c|}{}                          & \multicolumn{1}{c|}{}                         & \multicolumn{1}{c|}{PGD}      & \multicolumn{1}{c|}{45.52\%}          & \multicolumn{1}{c|}{32.11\%}                                  & \multicolumn{1}{c|}{30.82\%}                                  & \multicolumn{1}{c|}{\cellcolor[HTML]{C0C0C0}33.50\%}          & \multicolumn{1}{c|}{42.53\%}          & \multicolumn{1}{c|}{49.82\%}          & \multicolumn{1}{c|}{27.2361} & \multicolumn{1}{c|}{0.8032} & \multicolumn{1}{c|}{19.3294} & 0.1994             \\
\multicolumn{1}{c|}{}                          & \multicolumn{1}{c|}{}                         & \multicolumn{1}{c|}{Diff-PGD} & \multicolumn{1}{c|}{46.37\%}          & \multicolumn{1}{c|}{34.18\%}                                  & \multicolumn{1}{c|}{30.30\%}                                  & \multicolumn{1}{c|}{\cellcolor[HTML]{C0C0C0}\hl{59.60\%}} & \multicolumn{1}{c|}{45.33\%}          & \multicolumn{1}{c|}{52.29\%}          & \multicolumn{1}{c|}{27.1603} & \multicolumn{1}{c|}{0.8001} & \multicolumn{1}{c|}{20.1820} & 0.2024             \\
\multicolumn{1}{c|}{}                          & \multicolumn{1}{c|}{\multirow{-3}{*}{$p_1+$}} & \multicolumn{1}{c|}{Ours}     & \multicolumn{1}{c|}{\hl{52.58\%}} & \multicolumn{1}{c|}{\hl{42.90\%}}                         & \multicolumn{1}{c|}{\hl{39.50\%}}                         & \multicolumn{1}{c|}{\cellcolor[HTML]{C0C0C0}56.58\%}          & \multicolumn{1}{c|}{\hl{50.46\%}} & \multicolumn{1}{c|}{\hl{54.06\%}} & \multicolumn{1}{c|}{27.0858} & \multicolumn{1}{c|}{0.7954} & \multicolumn{1}{c|}{21.5762} & 0.2072             \\ \cline{2-13} 
\multicolumn{1}{c|}{}                          & \multicolumn{1}{c|}{}                         & \multicolumn{1}{c|}{PGD}      & \multicolumn{1}{c|}{62.74\%}          & \multicolumn{1}{c|}{50.16\%}                                  & \multicolumn{1}{c|}{52.11\%}                                  & \multicolumn{1}{c|}{\cellcolor[HTML]{C0C0C0}54.70\%}          & \multicolumn{1}{c|}{60.92\%}          & \multicolumn{1}{c|}{63.10\%}          & \multicolumn{1}{c|}{26.7231} & \multicolumn{1}{c|}{0.8121} & \multicolumn{1}{c|}{29.6638} & 0.2282             \\
\multicolumn{1}{c|}{}                          & \multicolumn{1}{c|}{}                         & \multicolumn{1}{c|}{Diff-PGD} & \multicolumn{1}{c|}{63.20\%}          & \multicolumn{1}{c|}{52.91\%}                                  & \multicolumn{1}{c|}{51.96\%}                                  & \multicolumn{1}{c|}{\cellcolor[HTML]{C0C0C0}81.50\%}          & \multicolumn{1}{c|}{65.27\%}          & \multicolumn{1}{c|}{65.40\%}          & \multicolumn{1}{c|}{26.5851} & \multicolumn{1}{c|}{0.8046} & \multicolumn{1}{c|}{32.5457} & 0.2408             \\
\multicolumn{1}{c|}{}                          & \multicolumn{1}{c|}{\multirow{-3}{*}{$p_2$}}  & \multicolumn{1}{c|}{Ours}     & \multicolumn{1}{c|}{\hl{79.79\%}} & \multicolumn{1}{c|}{\hl{73.59\%}}                         & \multicolumn{1}{c|}{\hl{73.28\%}}                         & \multicolumn{1}{c|}{\cellcolor[HTML]{C0C0C0}\hl{96.48\%}} & \multicolumn{1}{c|}{\hl{78.99\%}} & \multicolumn{1}{c|}{\hl{75.95\%}} & \multicolumn{1}{c|}{26.5331} & \multicolumn{1}{c|}{0.7990} & \multicolumn{1}{c|}{36.4909} & 0.2514             \\ \cline{2-13} 
\multicolumn{1}{c|}{}                          & \multicolumn{1}{c|}{}                         & \multicolumn{1}{c|}{PGD}      & \multicolumn{1}{c|}{63.44\%}          & \multicolumn{1}{c|}{55.11\%}                                  & \multicolumn{1}{c|}{54.77\%}                                  & \multicolumn{1}{c|}{\cellcolor[HTML]{C0C0C0}50.40\%}          & \multicolumn{1}{c|}{66.34\%}          & \multicolumn{1}{c|}{61.44\%}          & \multicolumn{1}{c|}{24.8592} & \multicolumn{1}{c|}{0.7740} & \multicolumn{1}{c|}{39.6970} & 0.2698             \\
\multicolumn{1}{c|}{}                          & \multicolumn{1}{c|}{}                         & \multicolumn{1}{c|}{Diff-PGD} & \multicolumn{1}{c|}{62.11\%}          & \multicolumn{1}{c|}{54.68\%}                                  & \multicolumn{1}{c|}{53.85\%}                                  & \multicolumn{1}{c|}{\cellcolor[HTML]{C0C0C0}63.20\%}          & \multicolumn{1}{c|}{69.74\%}          & \multicolumn{1}{c|}{61.59\%}          & \multicolumn{1}{c|}{24.7774} & \multicolumn{1}{c|}{0.7688} & \multicolumn{1}{c|}{42.3275} & 0.2779             \\
\multicolumn{1}{c|}{\multirow{-12}{*}{DN121}}  & \multicolumn{1}{c|}{\multirow{-3}{*}{$p_2+$}} & \multicolumn{1}{c|}{Ours}     & \multicolumn{1}{c|}{\hl{75.08\%}} & \multicolumn{1}{c|}{\hl{70.08\%}}                         & \multicolumn{1}{c|}{\hl{66.60\%}}                         & \multicolumn{1}{c|}{\cellcolor[HTML]{C0C0C0}\hl{88.84\%}} & \multicolumn{1}{c|}{\hl{78.37\%}} & \multicolumn{1}{c|}{\hl{70.65\%}} & \multicolumn{1}{c|}{24.7393} & \multicolumn{1}{c|}{0.7646} & \multicolumn{1}{c|}{46.0230} & 0.2857             \\ \hline
\end{tabular}%
}
\caption{Comparison of our proposed method with baseline methods in terms of Attack Success Rate (ASR) and Image Quality Assessment (IQA) metrics. $p_1$ refers to classifiers defended using the same setup as Diff-PGD, while $p_2$ denotes classifiers defended with adversarial purification as in our adaptive attack setup. $p_n+$ indicates scenarios where the adversary explicitly knows the input image is an adversarial sample, showing results after an additional purification step on top of the $p_n$-defended classifier. Cells with gray backgrounds represent white-box attack scenarios. Results are grouped in vertical sets of three, with the best result in each group highlighted in coral. $\uparrow$/$\downarrow$ indicate the higher/lower value has the better image quality.}
\label{tab:asr-iqa}
\end{table*}

\section{Experiments}
\subsection{Experimental Setup}
We compared our proposed method with
PGD~\cite{madry2017towards},
and
Diff-PGD~\cite{xue2024diffusion}, which the Diff-PGD is currently the most powerful attack.
We craft the adversarial samples on ResNet-50 (RN50)~\cite{he2016deep}, WideResNet-101 (WR101)~\cite{zagoruyko2016wide}, and DenseNet-121 (DN121)~\cite{huang2017densely}.
To assess the transferability and robustness of the crafted adversarial samples, we also tested the performance of some popular neural network architectures, including
VGG-19~\cite{simonyan2015very},
MobileNet v2~\cite{sandler2018mobilenetv2},
and
ShuffleNet v2~\cite{ma2018shufflenet},
with crafted adversarial samples.
We evaluate adversarial samples with a series of Image Quality Assessment (IQA): Peak Signal-to-Noise Ratio (PSNR), Structure Similarity Index Metric (SSIM)~\cite{wang2004image}, Fr\'echet Inception Distance (FID)~\cite{heusel2017gans}, Learned Perceptual Image Patch Similarity (LPIPS)~\cite{zhang2018unreasonable}. We used a subset of ImageNet~\cite{deng2009imagenet} proposed by~\citeauthor{lin2020nesterov}~\cite{lin2020nesterov}, which is a convention in adversarial attack to implement fast but representative experiments.

Comparing attack methods on undefended models has become less meaningful, as most state-of-the-art attack techniques can achieve high attack success rates in such scenarios. Therefore, we focus our evaluation on the performance of each method against defended classifiers. This approach provides a more realistic and challenging assessment of adversarial robustness.

For each method, we conducted tests under two defense scenarios
a) Standard defense: The classifier is trained with defensive techniques but does not know whether the input is an adversarial sample.
b) Enhanced defense: The classifier is aware that the input might be adversarial and employs an additional step of adversarial purification.
To achieve fair comparison, we set $\alpha=2/255$, $t=10$, $\epsilon=16/255$ for all methods, for the in-attack purification, we follow the setting of Diff-PGD, making the noising strength of MMA purification as $3/50$; relatively, the strength of MMAD-Purify is set as $0.05$. For the defense after attack loops, we set the strength of MMA and MMAD-Purify as $5/50$ and $0.1$.

\subsection{Results}

\begin{figure}[tb]
    \centering
    \includegraphics[width=\columnwidth]{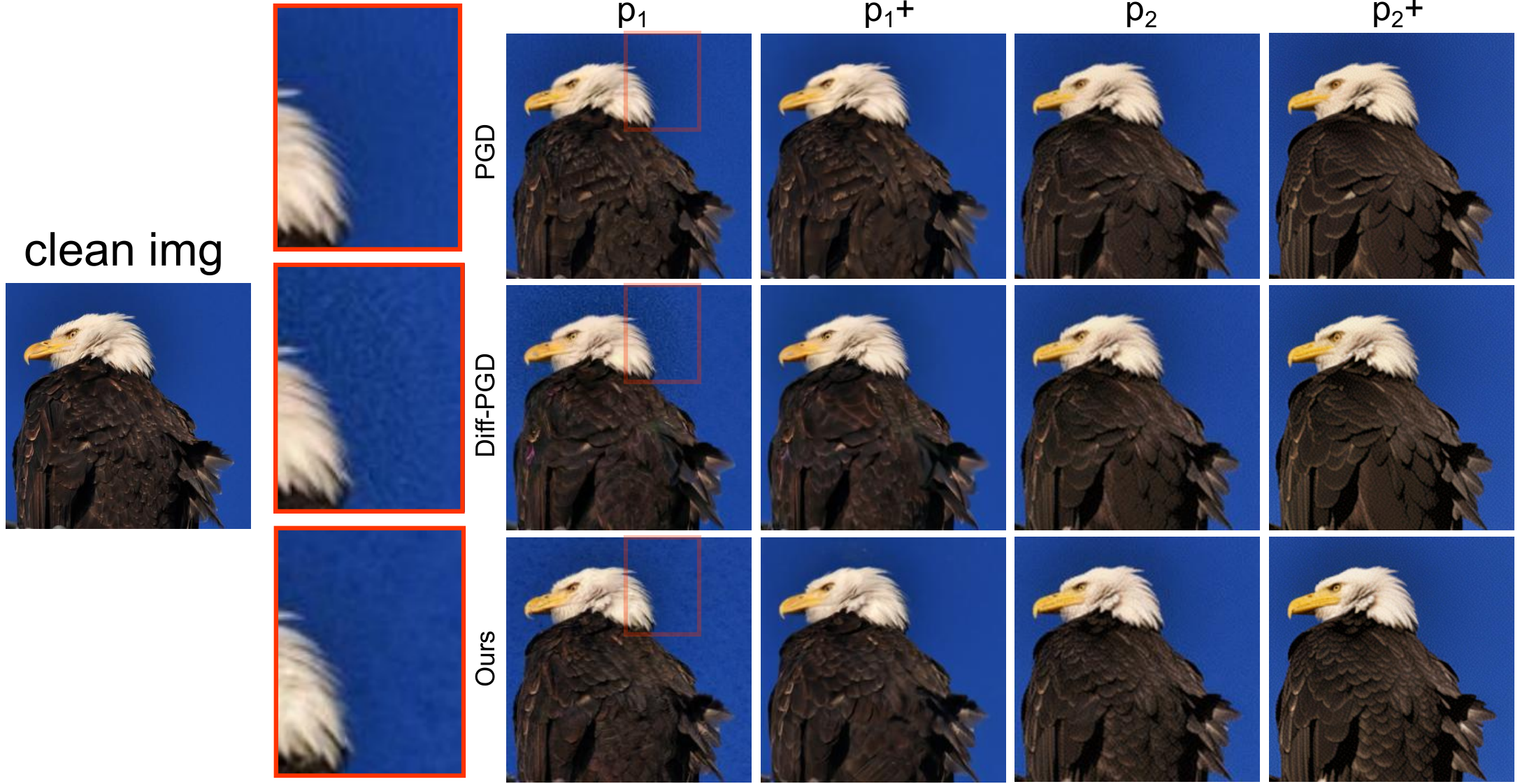}
    \caption{An example to visualize the difference of ResNet-50 generated adversarial samples between 2 different defense techniques across 3 methods. By zooming in on the left, the adversarial perturbation generated by our method is more imperceptible, and even when the defense method is a black box ($p_1$), the anti-purification capability is still strong.}
    \label{fig:cross-3-methods}
\end{figure}

As shown in Table~\ref{tab:asr-iqa}, our proposed method demonstrates overwhelming superiority compared to other baselines across almost all scenarios. The attack success rates (ASR) of our method consistently outperform those of PGD and Diff-PGD, particularly in more challenging defense setups ($p_n+$). For instance, under the $p_2$ defense scenario, our method achieves ASRs that are often 10-20 percentage points higher than the best baseline method.

Interestingly, in white-box scenarios under $p_1$ and $p_1+$ defenses, our method shows slightly lower ASRs compared to Diff-PGD. This phenomenon can be attributed to the fact that our method is designed to generate more transferable adversarial examples, which may come at the cost of slight performance degradation in very few scenarios. However, this trade-off is well compensated by the superior performance in most scenarios and more robust defense settings.

This performance advantage is maintained even in the most stringent defense scenario ($p_2+$), where an additional purification step is applied. Here, our method continues to exhibit superior attack capability, underlining its robustness against advanced defensive measures. The effectiveness of our approach is further emphasized by its performance across various network architectures, showcasing its versatility and generalizability.
Notably, the transferability of our attack method is evident from its high ASRs on models different from the one used to craft the adversarial samples. This cross-model performance indicates that our method generates adversarial perturbations that are more generalizable and less model-specific compared to existing methods.
Regarding image quality metrics, while our method shows slightly lower PSNR and SSIM values compared to baselines, the differences are marginal. The slight decrease in these metrics is a reasonable trade-off considering the significant gain in ASRs. Moreover, the FID and LPIPS scores for our method, while slightly higher, remain comparable to those of baseline methods, suggesting that the perceptual quality of our adversarial samples remains competitive.

\begin{figure}[ht]
    \centering
    \includegraphics[width=\linewidth]{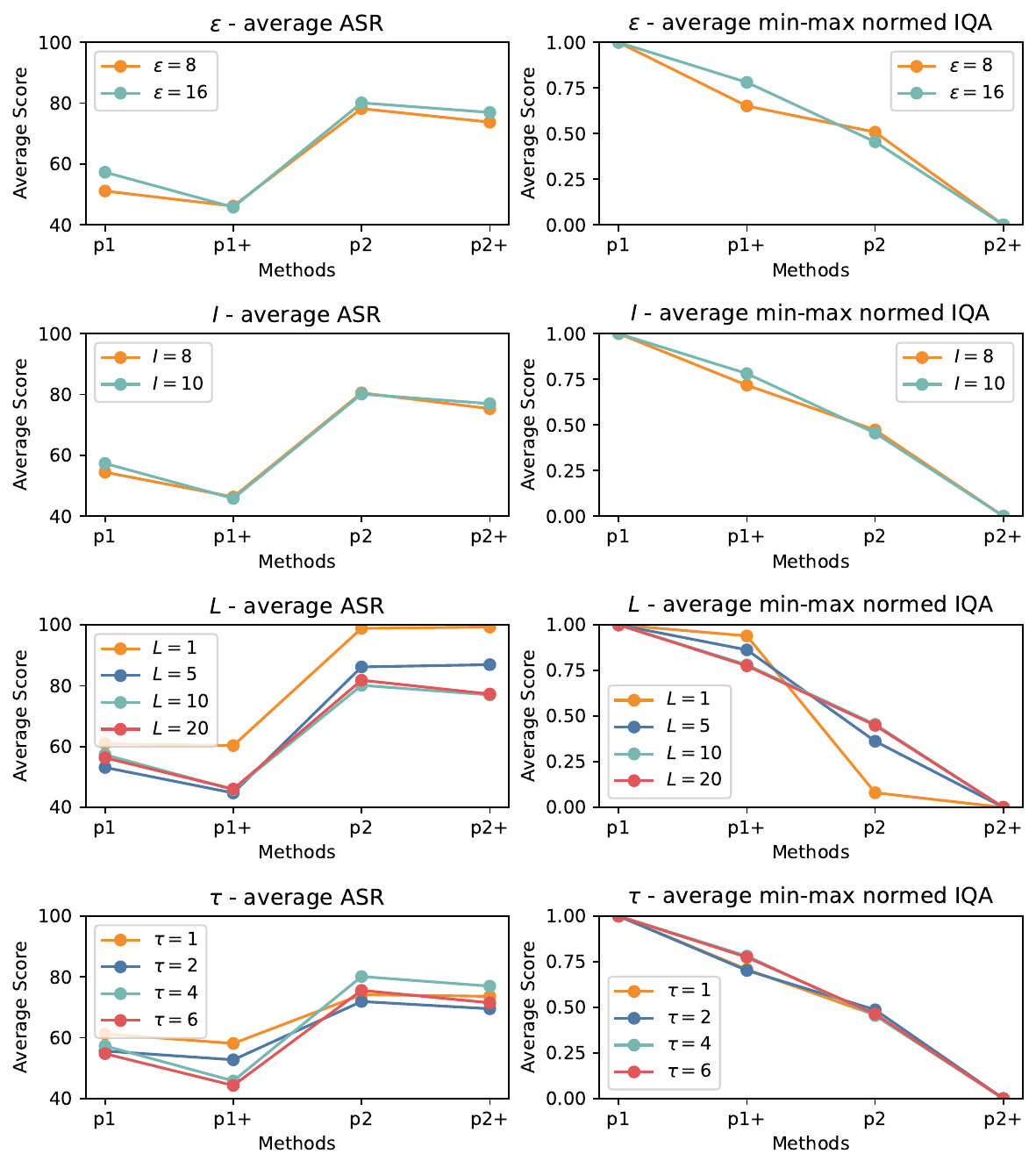}
    \caption{Ablation study on $\epsilon$, $I$, $L$, and $\tau$. Avg ASR is the mean performance across six neural networks. Avg IQA is calculated using min-max normalization of PSNR, SSIM, FID, and LPIPS, with 1-minmax applied to FID and LPIPS for consistency in quality representation.}
    \label{fig:ablation}
\end{figure}

\subsection{Ablation Study}
As shown in Figure~\ref{fig:ablation}, we conduct ablation studies on different hyperparameters to evaluate their influence
\subsubsection{Ablation on attack budget}
Experiments with $\epsilon=8/255$ and $\epsilon=16/255$ reveal that larger perturbation budgets produce stronger attacks but decrease image quality. Interestingly, under the $p_1+$ scenario, smaller budgets sometimes result in better transferability. We choose $\epsilon = 16/255$ for our primary experiments, balancing attack strength and image quality, allowing MMAD-Purify to demonstrate superior performance across various scenarios.
\subsubsection{Ablation on attack step}
We examined cases where $I = 8$ and $I=10$. Increasing attack steps improves convergence and transferability of adversarial examples. Choosing $I = 10$ enhances performance significantly. While this might increase computational time, the trade-off is mitigated by using distilled DM. This combination allows MMAD-Purify to maximize attack potential while maintaining efficiency, achieving a balance between effectiveness and practicality.
\subsubsection{Ablation on precision-optimized steps}
Our study on precision-optimized steps $L$ revealed a trade-off between ASR and image quality. Higher $L$ values enhance ASR but degrade image quality. We determined $L=10$ as optimal, balancing image integrity and attack effectiveness. This choice is crucial for MMAD-Purify in multi-modal scenarios, allowing the generation of adversarial examples that fool target models while maintaining visual similarity to originals. This fine-tuning solidifies MMAD-Purify's position as a state-of-the-art solution for multi-modal adversarial attacks.
\subsubsection{Ablation on timesteps of distilled diffusion model}
Our study on the timesteps $\tau$ of the distilled diffusion model revealed that both excessively large and small values lead to suboptimal performance, especially in white-box settings. We observed significant ASR fluctuations across all scenarios as $\tau$ varied, with minimal impact on IQA metrics. The model performed optimally when $\tau = 4$, balancing attack efficacy and computational efficiency. This finding was consistent across various conditions, highlighting the importance of careful parameter tuning. We selected $\tau = 4$ for our subsequent experiments, ensuring robust performance across diverse scenarios.This choice significantly benefits MMAD-Purify by enhancing its computational efficiency while maintaining high attack success rates. 

\section{Conclusion}
In this paper, we introduced MMAD-Purify, a novel multi-modality adversarial attack method designed to extend adversarial techniques across various data modalities. MMAD-Purify integrates a diffusion model and optimizes the precision of the noise predictor to enhance model robustness against adversarial attacks. Additionally, it improves the speed and transferability of adversarial samples through adversarial purification. Our experiments demonstrate the practical value of MMAD-Purify, making it a valuable tool for real-world applications. However, due to computational resource limitations, our experiments were focused on a limited set of modalities. In future work, we plan to extend our evaluation to additional modalities, such as depth maps and optical flow, to further explore the potential and versatility of MMAD-Purify in diverse scenarios. Overall, MMAD-Purify represents a significant advancement in multi-modality adversarial attacks, offering a robust and efficient solution to enhance model security in the face of evolving threats.
% %File: anonymous-submission-latex-2025.tex
% \documentclass[letterpaper]{article} % DO NOT CHANGE THIS
% \usepackage[submission]{aaai25}  % DO NOT CHANGE THIS
% \usepackage{times}  % DO NOT CHANGE THIS
% \usepackage{helvet}  % DO NOT CHANGE THIS
% \usepackage{courier}  % DO NOT CHANGE THIS
% \usepackage[hyphens]{url}  % DO NOT CHANGE THIS
% \usepackage{graphicx} % DO NOT CHANGE THIS
% \urlstyle{rm} % DO NOT CHANGE THIS
% \def\UrlFont{\rm}  % DO NOT CHANGE THIS
% \usepackage{natbib}  % DO NOT CHANGE THIS AND DO NOT ADD ANY OPTIONS TO IT
% \usepackage{caption} % DO NOT CHANGE THIS AND DO NOT ADD ANY OPTIONS TO IT
% \usepackage{amsmath}
% \usepackage{amssymb}
% \usepackage{multirow}
% \usepackage{graphicx}
% \usepackage{colortbl}
% \usepackage{xcolor}
% \usepackage{algorithm}
% \usepackage{algorithmic}
\definecolor{coral}{RGB}{255,127,80}
\frenchspacing  % DO NOT CHANGE THIS
\setlength{\pdfpagewidth}{8.5in} % DO NOT CHANGE THIS
\setlength{\pdfpageheight}{11in} % DO NOT CHANGE THIS
\title{Appendix for}

\onecolumn
\appendix
\section{\huge Appendix for ``MMAD-Purify: A Precision-Optimized Framework for Efficient and Scalable Multi-Modal Attacks'}
\vspace{3em}
% \subsection{Experiment Results for SDXL-PGD and SDXL-Turbo-PGD}

% To support the ``\emph{Renoise-PGD}'' of \textbf{Method} in main paper, we offer the comparison of SDXL-PGD and SDXL-Turbo-PGD.

% Table~\ref{tab:supp:sdxl:asr-iqa} illustrates the attack success rate (ASR) of both two SDXL-based methods. The adversarial samples are generated with surrogate model ResNet-50.
% All parameters are set equal to or equivalent to the Diff-PGD.

% SDXL-based methods perform poor ASR in the white-box scenario (53.59\% and 54.88\% v.s. 78.98\%), and we witness a strong image quality degrade after $p_1+$, the corresponding PSNR (27.2 $\rightarrow$ 22.6), SSIM (0.8 $\rightarrow$ 0.55), FID (19.5 $\rightarrow$ 39.5), and LPIPS (0.2 $\rightarrow$ 0.4) showcase how significant image quality degrade the SDXL-based methods suffered. This makes the perturbation obvious enough to be recognized by human.

% \begin{table*}[htb]
% \centering
% \input{supp/sdxl-asr-iqa}
% \caption{The comparision of SDXL-PGD and SDXL-Turbo-PGD. $p_1$ refers to classifiers defended using the same setup as diff-pgd, while $p_1+$ indicates scenarios where the adversary explicitly knows the input image is an adversarial sample, showing results after an additional purification step on top of the $p_1$-defended classifier. Cells with gray backgrounds represent white-box attack scenarios. Results are grouped in vertical sets of two, with the best image quality in each group highlighted in coral. $\uparrow$/$\downarrow$ indicate the higher/lower value has the better image quality.}
% \label{tab:supp:sdxl:asr-iqa}
% \end{table*}
This Appendix provides supplementary technical details to our submitted paper. The document is organized into the following sections:
\begin{itemize}
\item Detailed Comparison of MMA, MMAD, and MMAD-Purify
\item Technical Details of Distillation Methodology for Diffusion Models
\item Region-Specific Adversarial Attack (MMAD-Region)
\item Additional Results for Image Inpainting Tasks
\item Metadata for Ablation Studies Reported in the Main Paper
\end{itemize}
\subsection{Detailed Comparison of MMA, MMAD, and MMAD-Purify}

\begin{figure}[ht]
    \centering
    \includegraphics[width=0.7\linewidth]{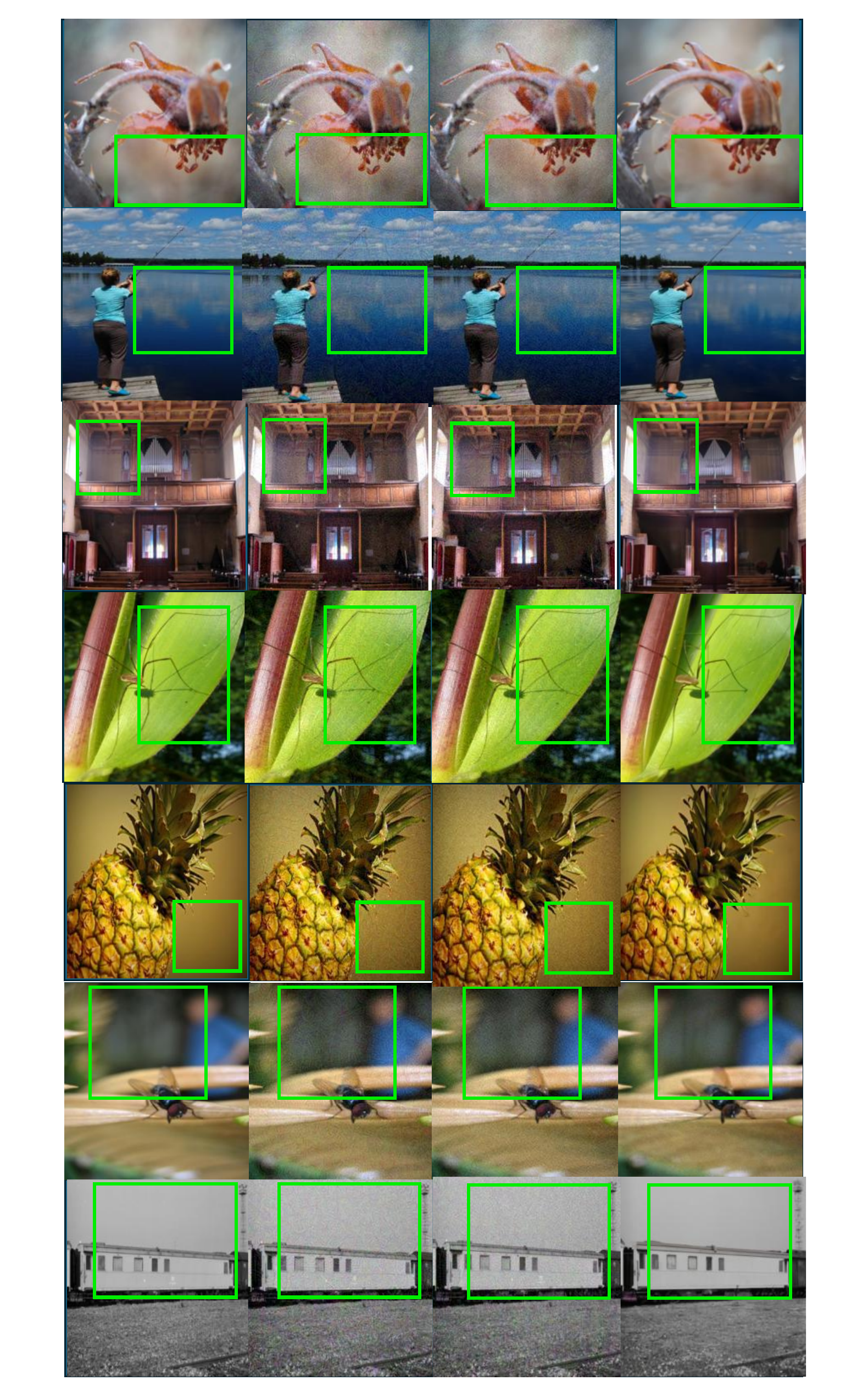}
    \caption{\textbf{Visualization Comparison of MMA, MMAD, and MMAD-Purify:} From left to right: original image $\mathbf{x}$, adversarial sample of MMA, adversarial sample of MMAD, and adversarial sample of MMAD-Purify.The green box highlights details of adversarial samples from different methods. We observe that demonstrates the highest fidelity to the original image compared to the other two attack methods, highlighting the effectiveness of our MMAD-Purify approach in preserving image quality while maintaining adversarial properties. In our experiments, we set the number of attack steps $I = 10$, attack budget $\epsilon = 16/255$, number of timesteps $\tau = 4$ for MMAD and MMAD-Purify, and $\tau = 50$ for MMA. The noising strengths for MMA, MMAD, and MMAD-Purify are $3/50$, $0.6$, and $0.05$, respectively, with the number of precision-optimized steps $L = 10$. In our initial experiments, we found that both the Image Quality Assessment (IQA) and Attack Success Rate (ASR) of MMAD were not ideal, which motivated the development of MMAD-Purify.}
    \label{fig:vs}
\end{figure}

In this section, we provide a comprehensive comparison of MMA, MMAD, and MMAD-Purify, focusing on their architectural differences and implementation details. The choice of backbone is crucial for the performance of each method. For MMA, we employ SDXL~\cite{podell2023sdxl} as the backbone, leveraging its advanced capabilities in image generation. In MMAD, we utilize LCM-LoRA~\cite{luo2023lcm} as our distilled diffusion model backbone. For MMAD-Purify, we opt for SDXL Turbo~\cite{sauer2023adversarial} as the distilled diffusion model backbone, capitalizing on its enhanced speed and quality. For a detailed explanation of the distilled diffusion model methodology, please refer to the section on distillation methodology for diffusion models~\ref{distillation}. This approach is fundamental to the efficiency improvements observed in MMAD and MMAD-Purify. We also provide a detailed algorithm for MMAD-Purify in Algorithm~\ref{alg:adversarial-attack}, which outlines the step-by-step process of our proposed method. For a visual comparison of the results produced by MMA, MMAD, and MMAD-Purify, please refer to Figure~\ref{fig:vs}.

\begin{algorithm}[tb]
\caption{MMAD-Purify}
\label{alg:adversarial-attack}
\begin{algorithmic}[1]
\REQUIRE Original image $\mathbf{x}$, Number of precision-optimized steps $L$, Sequence of timesteps $\{t_n\}$, Text condition $\mathbf{c}$, Classifier-Free Guidance Scale $\omega$, Noise schedule $\alpha(t), \sigma(t)$, Decoder $\mathcal{D}(\cdot)$, Iterations $I$, Step size $\eta$, Clip value $\epsilon$, Target classifier $h_{\phi}$, Loss function $\ell$, Number of timesteps $\tau$, A series of regularization weights $\left\{\alpha_l\right\}_{l=1}^{L}$
\STATE $\boldsymbol{\delta}^{(0)} = \mathbf{0}$
\STATE $\mathbf{z} = \mathcal{E}(\mathbf{x})$
\FOR{$i = 0$ \textbf{to} $I-1$}
    \STATE $\mathbf{z}_{t_{n-1}} \sim \mathcal{N}\left(\alpha(t_{n-1}) \mathbf{z}, \sigma^2(t_{n-1}) \mathbf{I}\right)$
    \FOR{$t = 0$ \textbf{to} $\tau$}
        \STATE $\mathbf{z}_{t_{n}}^{(0)} = \mathbf{z}_{t_{n-1}}$
        \STATE $\epsilon_{\theta}^{(0)} = \epsilon_\theta(\mathbf{z}_{t_{n}}^{(0)}, t_{n-1}, \mathbf{c}, \omega)$
        \STATE $\bar{\epsilon}^{(0)} = \epsilon_{\theta}^{(0)}$
        \FOR{$l = 0$ \textbf{to} $L-1$}
            \STATE $\epsilon_\theta^{(l)} = \epsilon_\theta(\mathbf{z}_{t_{n}}^{(l)}, t_{n}, \mathbf{c}, \omega)$
            \STATE $\epsilon_{\theta, \mathrm{reg}}^{(l)} = \text{Reg}(\epsilon_\theta^{(l)}, \alpha_l)$
            \STATE $\bar{\epsilon}^{(l)} = \text{Avg}(\epsilon_{\theta, \mathrm{reg}}^{(l)}, \bar{\epsilon}^{(l-1)})$
            \STATE $\mathbf{z}_{t_{n}}^{(l+1)} = \Psi(\mathbf{z}_{t_{n}}^{(l)}, t_n, t_{n-1}, \mathbf{c}, \epsilon_{\theta, \mathrm{reg}}^{(l)})$
        \ENDFOR
        \STATE $\bar{\epsilon}_{\theta, \mathrm{reg}}^{(L)} = \text{Reg}(\bar{\epsilon}_{\theta}^{(L)}, \alpha_l)$
        \STATE $\hat{\mathbf{z}}_{t_{n}} = \Psi\left(\mathbf{z}_{t_{n}}^{(L)}, t_n, t_{n-1}, \mathbf{c}, \bar{\epsilon}_{\theta, \text{reg}}^{(L)}\right)$
    \ENDFOR
    \RETURN $\hat{\mathbf{z}}_{\tau}$
    \STATE $\tilde{\mathbf{z}} \leftarrow \boldsymbol{f}_\theta\left(\hat{\mathbf{z}}_{\tau}, \omega, \mathbf{c}, \tau\right)$
    \STATE $\tilde{\mathbf{x}} = \mathcal{D}(\tilde{\mathbf{z}})$
    \STATE $\tilde{\boldsymbol{\delta}}^{(i)} = \boldsymbol{\delta}^{(i-1)} + \eta \nabla_{\boldsymbol{\delta}} \ell(h_{\phi}(\tilde{\mathbf{x}}, y))$
    \STATE $\boldsymbol{\delta}^{(i)} = \Pi_{\epsilon}(\tilde{\boldsymbol{\delta}}^{(i)})$
    \STATE $\mathbf{x}^{\text{adv}} = \mathbf{x} + \boldsymbol{\delta}^{(i)}$
\ENDFOR
\STATE $\mathbf{z}^{\text{adv}} = \mathcal{E}(\mathbf{x}^{\text{adv}})$
\STATE $\mathbf{z}_{t_n}^{\text{adv}} \sim \mathcal{N}\left(\alpha(t_n) \mathbf{z}^{\text{adv}}, \sigma^2(t_n) \mathbf{I}\right)$
\STATE $\mathbf{z}_{\text{adv}}^p = \text{MMAD-Purify}(\mathbf{z}_{t_n}^{\text{adv}})$
\STATE $\mathbf{x}_{\text{adv}}^p \leftarrow \mathcal{D}(\mathbf{z}_{\text{adv}}^p)$
\STATE \textbf{Output:} Robust adversarial sample $\mathbf{x}_{\text{adv}}^p$
\end{algorithmic}
\end{algorithm}

\begin{algorithm}[tb]
\caption{MMAD-Region}
\label{alg:adversarial-region}
\begin{algorithmic}[1]
\REQUIRE Original image $\mathbf{x}$, mask $M$, Number of precision-optimized steps $L$, Sequence of timesteps $\{t_n\}$, Text condition $\mathbf{c}$, Classifier-Free Guidance Scale $\omega$, Noise schedule $\alpha(t), \sigma(t)$, Decoder $\mathcal{D}(\cdot)$, Iterations $I$, Step size $\eta$, Clip value $\epsilon$, Target classifier $h_{\phi}$, Loss function $\ell$, Number of timesteps $\tau$, A series of regularization weights $\left\{\alpha_l\right\}_{l=1}^{L}$
\STATE $\mathbf{z} = \mathcal{E}(\mathbf{x})$
\FOR{$t = 0$ \textbf{to} $\tau$}
    \STATE $\mathbf{z}_{t_{n}}^{(0)} = \mathbf{z}_{t_{n-1}}$
    \STATE $\epsilon_{\theta}^{(0)} = \epsilon_\theta(\mathbf{z}_{t_{n}}^{(0)}, t_{n-1}, \mathbf{c}, \omega)$
    \STATE $\bar{\epsilon}^{(0)} = \epsilon_{\theta}^{(0)}$
    \FOR{$l = 0$ \textbf{to} $L-1$}
        \STATE $\epsilon_\theta^{(l)} = \epsilon_\theta(\mathbf{z}_{t_{n}}^{(l)}, t_{n}, \mathbf{c}, \omega)$
        \STATE $\epsilon_{\theta, \mathrm{reg}}^{(l)} = \text{Reg}(\epsilon_\theta^{(l)}, \alpha_l)$
        \STATE $\bar{\epsilon}^{(l)} = \text{Avg}(\epsilon_{\theta, \mathrm{reg}}^{(l)}, \bar{\epsilon}^{(l-1)})$
        \STATE $\mathbf{z}_{t_{n}}^{(l+1)} = \Psi(\mathbf{z}_{t_{n}}^{(l)}, t_n, t_{n-1}, \mathbf{c}, \epsilon_{\theta, \mathrm{reg}}^{(l)})$
    \ENDFOR
    \STATE $\bar{\epsilon}_{\theta, \mathrm{reg}}^{(L)} = \text{Reg}(\bar{\epsilon}_{\theta}^{(L)}, \alpha_L)$
    \STATE $\hat{\mathbf{z}}_{t_{n}} = \Psi\left(\mathbf{z}_{t_{n}}^{(L)}, t_n, t_{n-1}, \mathbf{c}, \bar{\epsilon}_{\theta, \text{reg}}^{(L)}\right)$
\ENDFOR
\STATE $\hat{\mathbf{z}}_{\tau} = \hat{\mathbf{z}}_{t_{n}}$
\STATE $\tilde{\mathbf{z}} \leftarrow \boldsymbol{f}_\theta\left(\hat{\mathbf{z}}_{\tau}, \omega, \mathbf{c}, \tau\right)$    
\STATE $\tilde{\mathbf{x}} = \mathcal{D}(\tilde{\mathbf{z}})$
\STATE $\bar{\epsilon}^{(0)} = \epsilon_{\theta}^{(0)}$
\FOR{$i = 0$ \textbf{to} $I-1$}
    \STATE $\tilde{\boldsymbol{\delta}}^{(i)} = \boldsymbol{\delta}^{(i-1)} + \eta \nabla_{\boldsymbol{\delta}} \ell(h_{\phi}(M \circ \tilde{\mathbf{x}}, y))$
    \STATE $\boldsymbol{\delta}^{(i)} = \Pi_{\epsilon}(\tilde{\boldsymbol{\delta}}^{(i)})$
    \STATE $\mathbf{x}^m_{\text{adv}} = M \circ \tilde{\mathbf{x}} + \boldsymbol{\delta}^{(i)}$
\ENDFOR
\STATE $\mathbf{x}^r_{\text{adv}} = \mathbf{x}^m_{\text{adv}} + (1-M) \circ \mathbf{x}$
\STATE \textbf{Output:} Robust adversarial sample for regional attack $\mathbf{x}^r_{\text{adv}}$
\end{algorithmic}
\end{algorithm}

\subsection{Distillation Methodology for Diffusion Models}
~\label{distillation}
In this section, we focus on the distillation methodology for diffusion models to better understand the principle of efficiency improvement for MMAD and MMAD-Purify. The following mathematical setup and methods are adapted from \cite{sauer2023adversarial}.
Let \(\mathcal{X}\) be the space of images. The training procedure involves an ADD-student \(\hat{x}_\theta: \mathcal{X} \times \mathbb{R}_{\geq 0} \to \mathcal{X}\) with parameters \(\theta \in \Theta\), a discriminator \(D_\phi: \mathcal{X} \to \mathbb{R}\) with parameters \(\phi \in \Phi\), and a DM teacher \(\hat{x}_\psi: \mathcal{X} \times \mathbb{R}_{\geq 0} \to \mathcal{X}\) with fixed parameters \(\psi \in \Psi\). The ADD-student generates samples \(\hat{x}_\theta(x_s, s)\) from noisy data \(x_s \in \mathcal{X}\), where \(x_s = \alpha_s x_0 + \sigma_s \epsilon\), \(x_0 \in \mathcal{X}\), \(\epsilon \sim \mathcal{N}(0, I)\) with \(\alpha_s, \sigma_s: \mathbb{R}_{\geq 0} \to \mathbb{R}\). Timesteps \(s\) are sampled uniformly from a finite set \(s \sim \mathcal{U}(T_{\text{student}})\), \(T_{\text{student}} = \{\tau_1, \ldots, \tau_n\} \subset \mathbb{R}_{\geq 0}\). The adversarial objective \(\mathcal{L}_{\text{adv}}^G: \mathcal{X} \times \Phi \to \mathbb{R}\) drives the discriminator to distinguish \(\hat{x}_\theta(x_s, s)\) from \(x_0\). The distillation loss \(\mathcal{L}_{\text{distill}}: \mathcal{X} \times \Psi \to \mathbb{R}\) uses the teacher's prediction \(\hat{x}_\psi(\hat{x}_{\theta, t}, t)\) as a target after further diffusing \(\hat{x}_\theta(x_s, s)\) to \(\hat{x}_{\theta, t}\).

The overall objective function \(\mathcal{L}: \Theta \times \Phi \to \mathbb{R}\) is defined as:
\begin{equation}
    \mathcal{L}(\theta, \phi) = \mathcal{L}_{\text{adv}}^G(\hat{x}_\theta(x_s, s), \phi) + \lambda \mathcal{L}_{\text{distill}}(\hat{x}_\theta(x_s, s), \psi),
\end{equation}
where \(\lambda \in \mathbb{R}_{>0}\) is a weighting factor.

By employing the distillation-based student model, our sampling process only uses 4 steps for both MMAD and MMAD-Purify. This approach significantly reduces the computational overhead while maintaining robust adversarial sample generation compared to MMA.
\clearpage

\subsection{Region-Specific Adversarial Attack}

In addition to global image manipulation, our MMAD framework demonstrates capabilities in region-specific adversarial attacks(MMAD-Region). This approach allows for targeted perturbations within designated areas of the image while preserving the integrity of the remaining content. By applying natural style transformations to specific regions, we achieve localized adversarial effects with minimal visual distortion.

Our region-specific attack methodology is characterized by its efficiency, requiring only a few attack iterations. In our experiments, we set the less number of attack iterations, significantly reducing computational overhead while maintaining attack efficacy. This streamlined process underscores the versatility and computational efficiency of MMAD-Region in various attack scenarios.

\begin{figure}[ht]
    \centering
    \includegraphics[width=\linewidth]{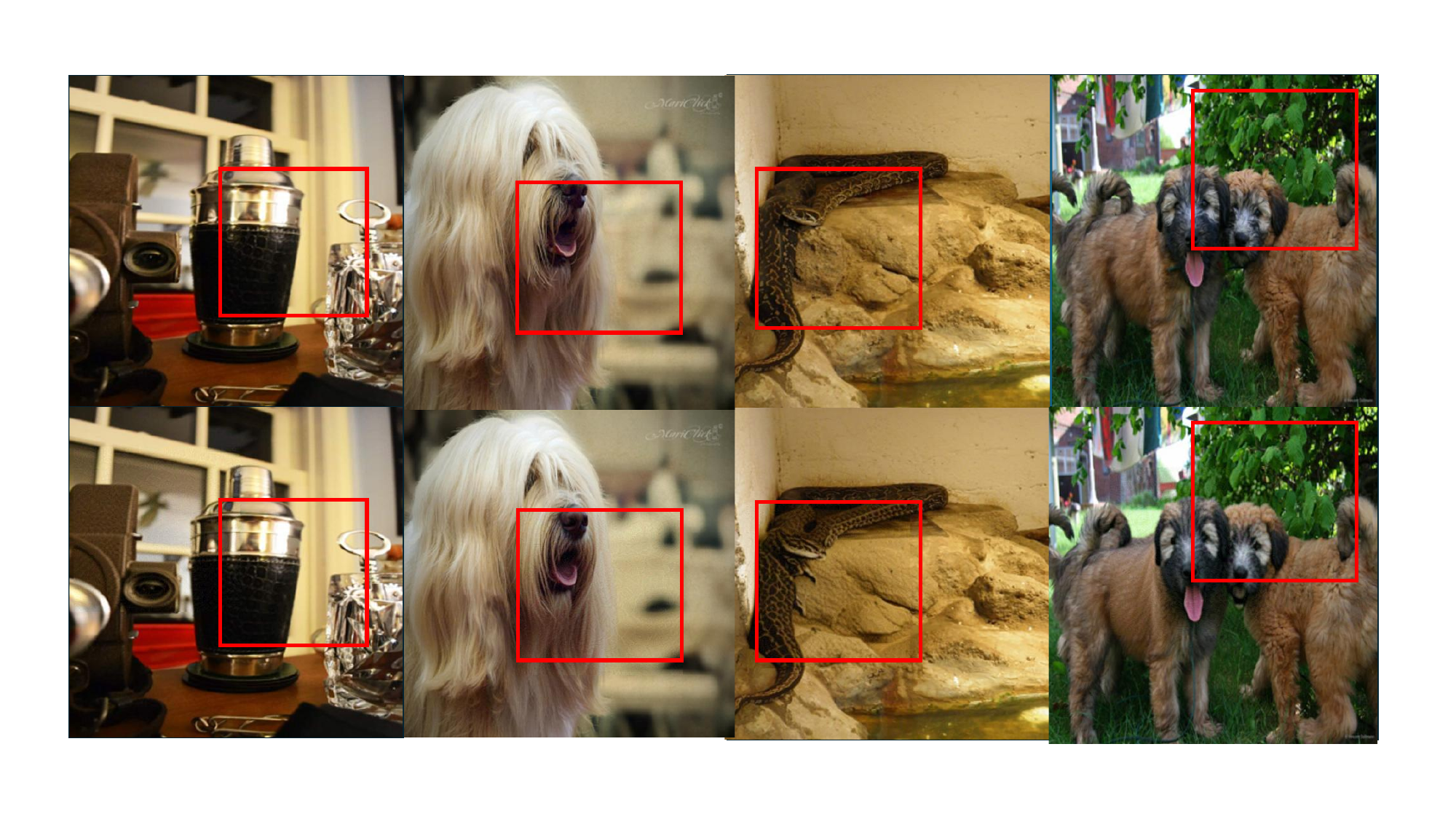}
    \caption{\textbf{Visualization of Adversarial Samples Generated by MMAD-Region:} MMAD-Region demonstrates the capability to produce highly realistic region-specific adversarial perturbations. The top row shows the original images, while the bottom row presents the corresponding adversarial samples generated by MMAD-Region. The attacked regions, demarcated by red bounding boxes, seamlessly integrate with the unmodified portions of the image, maintaining visual coherence and natural appearance. In our experiments, we set the number of attack steps $I = 3$, attack budget $\epsilon = 16/255$, number of timesteps $\tau = 4$, and number of precision-optimized steps $L = 10$.}
    \label{fig:region}
\end{figure}

The algorithmic implementation of our region-specific attack is delineated in Algorithm~\ref{alg:adversarial-region}, providing a comprehensive overview of the process. Figure~\ref{fig:region} showcases a diverse array of image examples, illustrating the effectiveness of our approach in generating regionally constrained adversarial perturbations while preserving the overall visual coherence of the image.

This region-specific capability extends the applicability of MMAD-Purify to scenarios requiring fine-grained control over adversarial manipulations, further demonstrating its flexibility and precision in adversarial attack generation.

\subsection{MMAD and MMAD-Purify for Image Inpainting Attacks }
\begin{figure}[ht]
    \centering
    \includegraphics[width=0.83\linewidth]{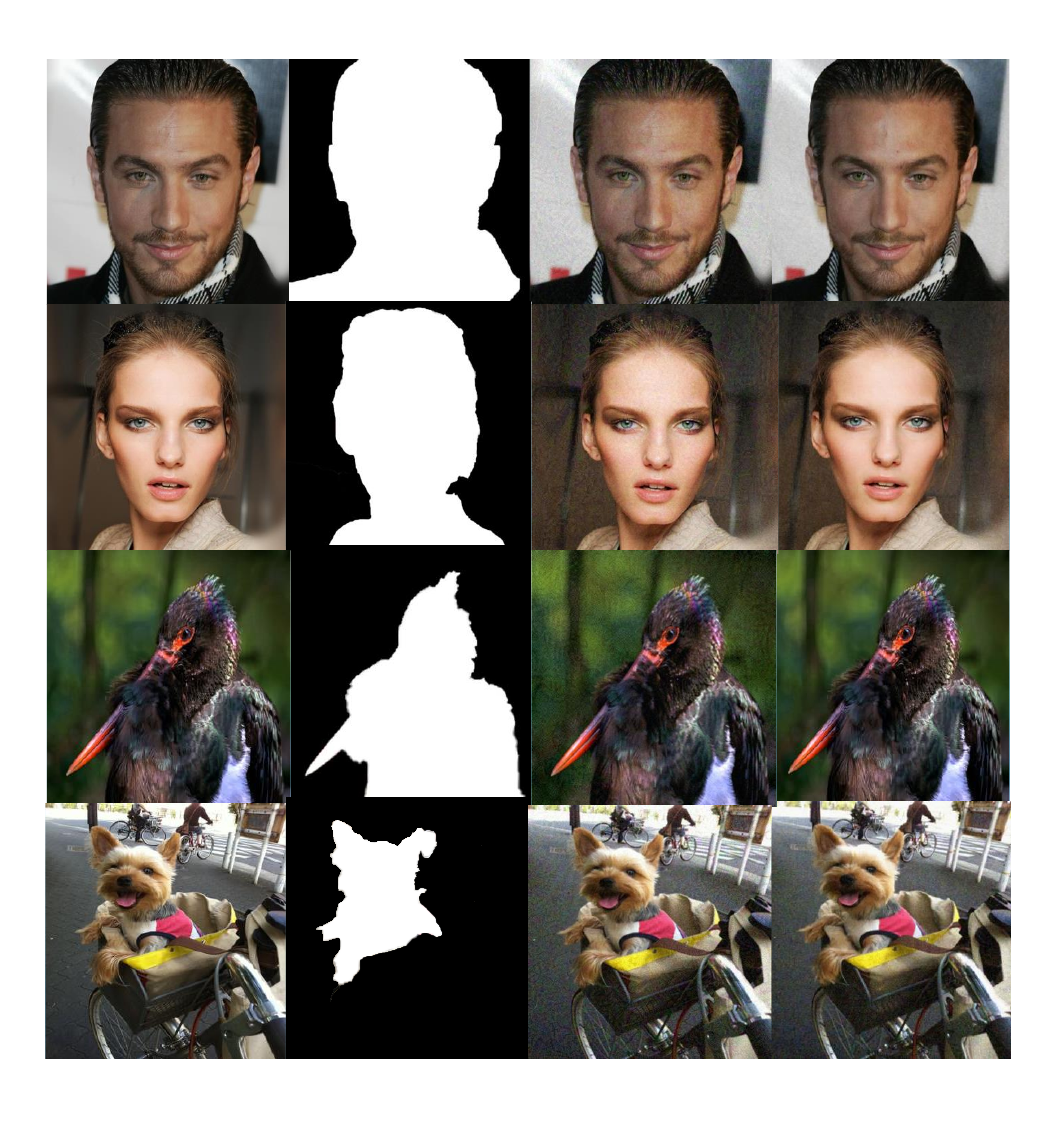}
    \caption{\textbf{Visualization of Adversarial Samples Generated by MMAD and MMAD-Purify for Inpainting Task:} From left to right: original image, segmentation mask, adversarial sample of MMAD, and adversarial sample  of MMAD-Purify. We can observe that MMAD-Purify produces results more visually similar to the original image than MMAD, as evidenced by the reduced presence of perceptible noise in the output.}
    \label{fig:inpainting}
\end{figure}

In our experiments on multi-modality adversarial attacks, we primarily focus on image-to-image transformations and inpainting attacks. This section provides supplementary results, specifically showcasing additional examples of our inpainting attack capabilities.
Our results demonstrate that the inpainting attack method is versatile and effective across a wide range of objectives. It successfully generates adversarial examples for common objects, such as dogs, as well as more complex targets like human faces. This broad applicability underscores the robustness and flexibility of our approach in diverse scenarios.
For these experiments, we maintained consistent hyperparameters across all trials to ensure comparability. Specifically, we set the number of attack steps $I = 10$, attack budget $\epsilon = 16/255$, number of timesteps $\tau = 4$, and number of precision-optimized steps $L = 10$. This configuration allowed us to balance attack effectiveness with computational efficiency while preserving image quality.
The Figure ~\ref{fig:inpainting} illustrate a range of inpainting attack results, demonstrating the method's ability to seamlessly integrate adversarial perturbations within the inpainted regions while maintaining visual coherence with the surrounding image content.
\clearpage
\subsection{Meta Data for Ablation Study}
Here we provide the meta data for ablation studies we reported in the main paper.
\vfill
% Please add the following required packages to your document preamble:
% \usepackage{multirow}
% \usepackage{graphicx}
% \usepackage[table,xcdraw]{xcolor}
% Beamer presentation requires \usepackage{colortbl} instead of \usepackage[table,xcdraw]{xcolor}
\begin{table}[ht]
\centering
% \resizebox{0.85\linewidth}{!}{%
\begin{tabular}{cccccccccccc}
\hline
\multicolumn{2}{c}{}                               & \multicolumn{6}{c}{neural networks}                                                                                                                                                                                                                                     & \multicolumn{4}{c}{IQAs}                                                                                                                  \\
\multicolumn{2}{c}{\multirow{-2}{*}{$L$}} & VGG19                                  & RN50                                                          & WR101                                 & DN121                                 & NNv2                                   & SNv2                                  & PSNR $\uparrow$                       & SSIM $\uparrow$                      & FID $\downarrow$                      & LPIPS $\downarrow$ \\ \hline
                         & \multicolumn{1}{c|}{1}  & \multicolumn{1}{c|}{\hl{68.80\%}}  & \multicolumn{1}{c|}{\cellcolor[HTML]{C0C0C0}62.11\%}          & \multicolumn{1}{c|}{\hl{55.05\%}} & \multicolumn{1}{c|}{\hl{62.50\%}} & \multicolumn{1}{c|}{\hl{59.51\%}}  & \multicolumn{1}{c|}{57.73\%}          & \multicolumn{1}{c|}{\hl{28.4538}} & \multicolumn{1}{c|}{\hl{0.8313}} & \multicolumn{1}{c|}{\hl{13.7087}} & \hl{0.1774}    \\
                         & \multicolumn{1}{c|}{5}  & \multicolumn{1}{c|}{60.36\%}           & \multicolumn{1}{c|}{\cellcolor[HTML]{C0C0C0}52.99\%}          & \multicolumn{1}{c|}{40.17\%}          & \multicolumn{1}{c|}{47.25\%}          & \multicolumn{1}{c|}{57.77\%}           & \multicolumn{1}{c|}{60.23\%}          & \multicolumn{1}{c|}{28.367}           & \multicolumn{1}{c|}{0.8266}          & \multicolumn{1}{c|}{17.6228}          & 0.1877             \\
                         & \multicolumn{1}{c|}{10} & \multicolumn{1}{c|}{58.76\%}           & \multicolumn{1}{c|}{\cellcolor[HTML]{C0C0C0}\hl{69.50\%}} & \multicolumn{1}{c|}{46.67\%}          & \multicolumn{1}{c|}{49.77\%}          & \multicolumn{1}{c|}{58.80\%}           & \multicolumn{1}{c|}{\hl{60.45\%}} & \multicolumn{1}{c|}{28.2141}          & \multicolumn{1}{c|}{0.8194}          & \multicolumn{1}{c|}{19.1739}          & 0.1964             \\
\multirow{-4}{*}{$p_1$}  & \multicolumn{1}{c|}{20} & \multicolumn{1}{c|}{57.64\%}           & \multicolumn{1}{c|}{\cellcolor[HTML]{C0C0C0}67.45\%}          & \multicolumn{1}{c|}{44.93\%}          & \multicolumn{1}{c|}{52.49\%}          & \multicolumn{1}{c|}{55.84\%}           & \multicolumn{1}{c|}{59.51\%}          & \multicolumn{1}{c|}{28.2208}          & \multicolumn{1}{c|}{0.8199}          & \multicolumn{1}{c|}{19.2446}          & 0.1957             \\ \hline
                         & \multicolumn{1}{c|}{1}  & \multicolumn{1}{c|}{\hl{66.00\%}}  & \multicolumn{1}{c|}{\cellcolor[HTML]{C0C0C0}\hl{57.14\%}} & \multicolumn{1}{c|}{\hl{53.21\%}} & \multicolumn{1}{c|}{\hl{59.52\%}} & \multicolumn{1}{c|}{\hl{65.59\%}}  & \multicolumn{1}{c|}{\hl{60.48\%}} & \multicolumn{1}{c|}{\hl{27.2092}} & \multicolumn{1}{c|}{\hl{0.8017}} & \multicolumn{1}{c|}{\hl{18.1056}} & \hl{0.1993}    \\
                         & \multicolumn{1}{c|}{5}  & \multicolumn{1}{c|}{53.39\%}           & \multicolumn{1}{c|}{\cellcolor[HTML]{C0C0C0}36.99\%}          & \multicolumn{1}{c|}{33.55\%}          & \multicolumn{1}{c|}{42.10\%}          & \multicolumn{1}{c|}{50.18\%}           & \multicolumn{1}{c|}{52.20\%}          & \multicolumn{1}{c|}{27.1812}          & \multicolumn{1}{c|}{0.7999}          & \multicolumn{1}{c|}{20.5338}          & 0.2032             \\
                         & \multicolumn{1}{c|}{10} & \multicolumn{1}{c|}{49.68\%}           & \multicolumn{1}{c|}{\cellcolor[HTML]{C0C0C0}44.52\%}          & \multicolumn{1}{c|}{36.04\%}          & \multicolumn{1}{c|}{41.27\%}          & \multicolumn{1}{c|}{50.08\%}           & \multicolumn{1}{c|}{53.31\%}          & \multicolumn{1}{c|}{27.1135}          & \multicolumn{1}{c|}{0.7968}          & \multicolumn{1}{c|}{20.9897}          & 0.2049             \\
\multirow{-4}{*}{$p_1+$} & \multicolumn{1}{c|}{20} & \multicolumn{1}{c|}{51.23\%}           & \multicolumn{1}{c|}{\cellcolor[HTML]{C0C0C0}45.71\%}          & \multicolumn{1}{c|}{35.51\%}          & \multicolumn{1}{c|}{40.87\%}          & \multicolumn{1}{c|}{49.92\%}           & \multicolumn{1}{c|}{52.88\%}          & \multicolumn{1}{c|}{27.1109}          & \multicolumn{1}{c|}{0.7967}          & \multicolumn{1}{c|}{21.2021}          & 0.2055             \\ \hline
                         & \multicolumn{1}{c|}{1}  & \multicolumn{1}{c|}{\hl{98.40\%}}  & \multicolumn{1}{c|}{\cellcolor[HTML]{C0C0C0}\hl{98.76\%}} & \multicolumn{1}{c|}{\hl{98.17\%}} & \multicolumn{1}{c|}{\hl{98.21\%}} & \multicolumn{1}{c|}{\hl{100.00\%}} & \multicolumn{1}{c|}{\hl{99.66\%}} & \multicolumn{1}{c|}{14.9437}          & \multicolumn{1}{c|}{0.4193}          & \multicolumn{1}{c|}{109.5199}         & 0.5795             \\
                         & \multicolumn{1}{c|}{5}  & \multicolumn{1}{c|}{88.57\%}           & \multicolumn{1}{c|}{\cellcolor[HTML]{C0C0C0}90.19\%}          & \multicolumn{1}{c|}{83.12\%}          & \multicolumn{1}{c|}{82.77\%}          & \multicolumn{1}{c|}{88.87\%}           & \multicolumn{1}{c|}{83.56\%}          & \multicolumn{1}{c|}{24.286}           & \multicolumn{1}{c|}{0.7541}          & \multicolumn{1}{c|}{51.1812}          & 0.2996             \\
                         & \multicolumn{1}{c|}{10} & \multicolumn{1}{c|}{80.25\%}           & \multicolumn{1}{c|}{\cellcolor[HTML]{C0C0C0}95.50\%}          & \multicolumn{1}{c|}{75.68\%}          & \multicolumn{1}{c|}{73.60\%}          & \multicolumn{1}{c|}{80.13\%}           & \multicolumn{1}{c|}{75.61\%}          & \multicolumn{1}{c|}{\hl{26.4921}} & \multicolumn{1}{c|}{\hl{0.797}}  & \multicolumn{1}{c|}{\hl{36.2468}} & \hl{0.2549}    \\
\multirow{-4}{*}{$p_2$}  & \multicolumn{1}{c|}{20} & \multicolumn{1}{c|}{83.09\%}           & \multicolumn{1}{c|}{\cellcolor[HTML]{C0C0C0}96.22\%}          & \multicolumn{1}{c|}{74.64\%}          & \multicolumn{1}{c|}{76.47\%}          & \multicolumn{1}{c|}{84.32\%}           & \multicolumn{1}{c|}{76.09\%}          & \multicolumn{1}{c|}{26.4823}          & \multicolumn{1}{c|}{0.7959}          & \multicolumn{1}{c|}{36.3049}          & 0.2562             \\ \hline
                         & \multicolumn{1}{c|}{1}  & \multicolumn{1}{c|}{\hl{100.00\%}} & \multicolumn{1}{c|}{\cellcolor[HTML]{C0C0C0}\hl{99.38\%}} & \multicolumn{1}{c|}{\hl{97.25\%}} & \multicolumn{1}{c|}{\hl{99.40\%}} & \multicolumn{1}{c|}{\hl{100.00\%}} & \multicolumn{1}{c|}{\hl{99.66\%}} & \multicolumn{1}{c|}{13.2084}          & \multicolumn{1}{c|}{0.369}           & \multicolumn{1}{c|}{112.1609}         & 0.6103             \\
                         & \multicolumn{1}{c|}{5}  & \multicolumn{1}{c|}{90.00\%}           & \multicolumn{1}{c|}{\cellcolor[HTML]{C0C0C0}86.46\%}          & \multicolumn{1}{c|}{85.68\%}          & \multicolumn{1}{c|}{83.84\%}          & \multicolumn{1}{c|}{90.81\%}           & \multicolumn{1}{c|}{84.70\%}          & \multicolumn{1}{c|}{22.1126}          & \multicolumn{1}{c|}{0.6987}          & \multicolumn{1}{c|}{67.3528}          & 0.3582             \\
                         & \multicolumn{1}{c|}{10} & \multicolumn{1}{c|}{76.27\%}           & \multicolumn{1}{c|}{\cellcolor[HTML]{C0C0C0}88.95\%}          & \multicolumn{1}{c|}{72.97\%}          & \multicolumn{1}{c|}{70.71\%}          & \multicolumn{1}{c|}{79.81\%}           & \multicolumn{1}{c|}{73.17\%}          & \multicolumn{1}{c|}{24.7144}          & \multicolumn{1}{c|}{\hl{0.7628}} & \multicolumn{1}{c|}{\hl{44.6224}} & \hl{0.2906}    \\
\multirow{-4}{*}{$p_2+$} & \multicolumn{1}{c|}{20} & \multicolumn{1}{c|}{76.68\%}           & \multicolumn{1}{c|}{\cellcolor[HTML]{C0C0C0}89.18\%}          & \multicolumn{1}{c|}{73.01\%}          & \multicolumn{1}{c|}{71.34\%}          & \multicolumn{1}{c|}{81.12\%}           & \multicolumn{1}{c|}{72.25\%}          & \multicolumn{1}{c|}{\hl{24.7179}} & \multicolumn{1}{c|}{0.7624}          & \multicolumn{1}{c|}{45.1594}          & 0.2916             \\ \hline
\end{tabular}%
% }
\caption{The ablation study for precision-optimized steps.}
\label{tab:abl:inferT}
\end{table}
\vfill
\begin{table}[ht]
\centering
% \resizebox{\linewidth}{!}{%
\begin{tabular}{cccccccccccc}
\hline
\multicolumn{2}{c}{}                               & \multicolumn{6}{c}{neural networks}                                                                                                                                                                                                                                   & \multicolumn{4}{c}{IQAs}                                                                                       \\
\multicolumn{2}{c}{\multirow{-2}{*}{$\tau$}} & VGG19                                 & RN50                                                          & WR101                                 & DN121                                 & NNv2                                  & SNv2                                  & PSNR $\uparrow$              & SSIM $\uparrow$             & FID $\downarrow$             & LPIPS $\downarrow$ \\ \hline
                          & \multicolumn{1}{c|}{1} & \multicolumn{1}{c|}{\hl{66.28\%}} & \multicolumn{1}{c|}{\cellcolor[HTML]{C0C0C0}61.18\%}          & \multicolumn{1}{c|}{\hl{51.26\%}} & \multicolumn{1}{c|}{\hl{68.48\%}} & \multicolumn{1}{c|}{\hl{62.61\%}} & \multicolumn{1}{c|}{57.89\%}          & \multicolumn{1}{c|}{28.44}   & \multicolumn{1}{c|}{0.8312} & \multicolumn{1}{c|}{13.3816} & 0.1758             \\
                          & \multicolumn{1}{c|}{2} & \multicolumn{1}{c|}{58.93\%}          & \multicolumn{1}{c|}{\cellcolor[HTML]{C0C0C0}47.28\%}          & \multicolumn{1}{c|}{50.82\%}          & \multicolumn{1}{c|}{59.38\%}          & \multicolumn{1}{c|}{58.26\%}          & \multicolumn{1}{c|}{59.21\%}          & \multicolumn{1}{c|}{28.4669} & \multicolumn{1}{c|}{0.8315} & \multicolumn{1}{c|}{14.8572} & 0.1797             \\
                          & \multicolumn{1}{c|}{4} & \multicolumn{1}{c|}{58.76\%}          & \multicolumn{1}{c|}{\cellcolor[HTML]{C0C0C0}\hl{69.50\%}} & \multicolumn{1}{c|}{46.67\%}          & \multicolumn{1}{c|}{49.77\%}          & \multicolumn{1}{c|}{58.80\%}          & \multicolumn{1}{c|}{\hl{60.45\%}} & \multicolumn{1}{c|}{28.2141} & \multicolumn{1}{c|}{0.8194} & \multicolumn{1}{c|}{19.1739} & 0.1964             \\
\multirow{-4}{*}{$p_1$}   & \multicolumn{1}{c|}{6} & \multicolumn{1}{c|}{56.60\%}          & \multicolumn{1}{c|}{\cellcolor[HTML]{C0C0C0}69.08\%}          & \multicolumn{1}{c|}{42.11\%}          & \multicolumn{1}{c|}{47.72\%}          & \multicolumn{1}{c|}{54.34\%}          & \multicolumn{1}{c|}{58.77\%}          & \multicolumn{1}{c|}{28.2464} & \multicolumn{1}{c|}{0.8214} & \multicolumn{1}{c|}{18.9801} & 0.1936             \\ \hline
                          & \multicolumn{1}{c|}{1} & \multicolumn{1}{c|}{\hl{63.95\%}} & \multicolumn{1}{c|}{\cellcolor[HTML]{C0C0C0}\hl{58.55\%}} & \multicolumn{1}{c|}{41.18\%}          & \multicolumn{1}{c|}{\hl{64.85\%}} & \multicolumn{1}{c|}{\hl{60.08\%}} & \multicolumn{1}{c|}{\hl{60.20\%}} & \multicolumn{1}{c|}{27.2103} & \multicolumn{1}{c|}{0.8019} & \multicolumn{1}{c|}{17.4615} & 0.1981             \\
                          & \multicolumn{1}{c|}{2} & \multicolumn{1}{c|}{58.63\%}          & \multicolumn{1}{c|}{\cellcolor[HTML]{C0C0C0}44.02\%}          & \multicolumn{1}{c|}{\hl{50.82\%}} & \multicolumn{1}{c|}{51.17\%}          & \multicolumn{1}{c|}{55.86\%}          & \multicolumn{1}{c|}{56.05\%}          & \multicolumn{1}{c|}{27.2187} & \multicolumn{1}{c|}{0.8014} & \multicolumn{1}{c|}{18.6287} & 0.2008             \\
                          & \multicolumn{1}{c|}{4} & \multicolumn{1}{c|}{49.68\%}          & \multicolumn{1}{c|}{\cellcolor[HTML]{C0C0C0}44.52\%}          & \multicolumn{1}{c|}{36.04\%}          & \multicolumn{1}{c|}{41.27\%}          & \multicolumn{1}{c|}{50.08\%}          & \multicolumn{1}{c|}{53.31\%}          & \multicolumn{1}{c|}{27.1135} & \multicolumn{1}{c|}{0.7968} & \multicolumn{1}{c|}{20.9897} & 0.2049             \\
\multirow{-4}{*}{$p_1+$}  & \multicolumn{1}{c|}{6} & \multicolumn{1}{c|}{50.15\%}          & \multicolumn{1}{c|}{\cellcolor[HTML]{C0C0C0}43.71\%}          & \multicolumn{1}{c|}{31.71\%}          & \multicolumn{1}{c|}{39.60\%}          & \multicolumn{1}{c|}{48.25\%}          & \multicolumn{1}{c|}{52.47\%}          & \multicolumn{1}{c|}{27.126}  & \multicolumn{1}{c|}{0.7975} & \multicolumn{1}{c|}{20.6414} & 0.2037             \\ \hline
                          & \multicolumn{1}{c|}{1} & \multicolumn{1}{c|}{78.29\%}          & \multicolumn{1}{c|}{\cellcolor[HTML]{C0C0C0}72.37\%}          & \multicolumn{1}{c|}{67.23\%}          & \multicolumn{1}{c|}{69.70\%}          & \multicolumn{1}{c|}{\hl{81.09\%}} & \multicolumn{1}{c|}{\hl{76.32\%}} & \multicolumn{1}{c|}{26.5986} & \multicolumn{1}{c|}{0.8032} & \multicolumn{1}{c|}{29.1863} & 0.2411             \\
                          & \multicolumn{1}{c|}{2} & \multicolumn{1}{c|}{76.19\%}          & \multicolumn{1}{c|}{\cellcolor[HTML]{C0C0C0}67.66\%}          & \multicolumn{1}{c|}{66.67\%}          & \multicolumn{1}{c|}{70.70\%}          & \multicolumn{1}{c|}{75.08\%}          & \multicolumn{1}{c|}{75.26\%}          & \multicolumn{1}{c|}{26.6929} & \multicolumn{1}{c|}{0.8074} & \multicolumn{1}{c|}{30.2083} & 0.2377             \\
                          & \multicolumn{1}{c|}{4} & \multicolumn{1}{c|}{\hl{80.25\%}} & \multicolumn{1}{c|}{\cellcolor[HTML]{C0C0C0}\hl{95.50\%}} & \multicolumn{1}{c|}{\hl{75.68\%}} & \multicolumn{1}{c|}{\hl{73.60\%}} & \multicolumn{1}{c|}{80.13\%}          & \multicolumn{1}{c|}{75.61\%}          & \multicolumn{1}{c|}{26.4921} & \multicolumn{1}{c|}{0.797}  & \multicolumn{1}{c|}{36.2468} & 0.2549             \\
\multirow{-4}{*}{$p_2$}   & \multicolumn{1}{c|}{6} & \multicolumn{1}{c|}{75.46\%}          & \multicolumn{1}{c|}{\cellcolor[HTML]{C0C0C0}91.54\%}          & \multicolumn{1}{c|}{66.95\%}          & \multicolumn{1}{c|}{68.23\%}          & \multicolumn{1}{c|}{77.93\%}          & \multicolumn{1}{c|}{73.42\%}          & \multicolumn{1}{c|}{26.5005} & \multicolumn{1}{c|}{0.798}  & \multicolumn{1}{c|}{35.1451} & 0.2523             \\ \hline
                          & \multicolumn{1}{c|}{1} & \multicolumn{1}{c|}{73.26\%}          & \multicolumn{1}{c|}{\cellcolor[HTML]{C0C0C0}72.37\%}          & \multicolumn{1}{c|}{67.23\%}          & \multicolumn{1}{c|}{\hl{75.76\%}} & \multicolumn{1}{c|}{\hl{81.09\%}} & \multicolumn{1}{c|}{72.04\%}          & \multicolumn{1}{c|}{24.7843} & \multicolumn{1}{c|}{0.7682} & \multicolumn{1}{c|}{39.6818} & 0.2812             \\
                          & \multicolumn{1}{c|}{2} & \multicolumn{1}{c|}{73.51\%}          & \multicolumn{1}{c|}{\cellcolor[HTML]{C0C0C0}61.14\%}          & \multicolumn{1}{c|}{64.48\%}          & \multicolumn{1}{c|}{69.14\%}          & \multicolumn{1}{c|}{80.78\%}          & \multicolumn{1}{c|}{68.16\%}          & \multicolumn{1}{c|}{24.8393} & \multicolumn{1}{c|}{0.7702} & \multicolumn{1}{c|}{40.8266} & 0.2797             \\
                          & \multicolumn{1}{c|}{4} & \multicolumn{1}{c|}{\hl{76.27\%}} & \multicolumn{1}{c|}{\cellcolor[HTML]{C0C0C0}\hl{88.95\%}} & \multicolumn{1}{c|}{\hl{72.97\%}} & \multicolumn{1}{c|}{70.71\%}          & \multicolumn{1}{c|}{79.81\%}          & \multicolumn{1}{c|}{\hl{73.17\%}} & \multicolumn{1}{c|}{24.7144} & \multicolumn{1}{c|}{0.7628} & \multicolumn{1}{c|}{44.6224} & 0.2906             \\
\multirow{-4}{*}{$p_2+$}  & \multicolumn{1}{c|}{6} & \multicolumn{1}{c|}{73.31\%}          & \multicolumn{1}{c|}{\cellcolor[HTML]{C0C0C0}78.95\%}          & \multicolumn{1}{c|}{65.27\%}          & \multicolumn{1}{c|}{65.10\%}          & \multicolumn{1}{c|}{78.08\%}          & \multicolumn{1}{c|}{68.31\%}          & \multicolumn{1}{c|}{24.7084} & \multicolumn{1}{c|}{0.7636} & \multicolumn{1}{c|}{44.5897} & 0.2889             \\ \hline
\end{tabular}%
% }
\caption{The ablation study for }
\label{tab:my-table}
\end{table}
\vfill
\clearpage

% Please add the following required packages to your document preamble:
% \usepackage{multirow}
% \usepackage{graphicx}
% \usepackage[table,xcdraw]{xcolor}
% Beamer presentation requires \usepackage{colortbl} instead of \usepackage[table,xcdraw]{xcolor}
\begin{table}[ht]
\centering
% \resizebox{\linewidth}{!}{%
\begin{tabular}{cccccccccccc}
\hline
\multicolumn{2}{c}{}                                                    & \multicolumn{6}{c}{neural networks}                                                                                                                                                                                                                                   & \multicolumn{4}{c}{IQAs}                                                                                       \\
\multicolumn{2}{c}{\multirow{-2}{*}{$\epsilon$}}                        & VGG19                                 & RN50                                                          & WR101                                 & DN121                                 & NNv2                                  & SNv2                                  & PSNR $\uparrow$              & SSIM $\uparrow$             & FID $\downarrow$             & LPIPS $\downarrow$ \\ \hline
\multicolumn{1}{c|}{}                         & \multicolumn{1}{c|}{8}  & \multicolumn{1}{c|}{54.83\%}          & \multicolumn{1}{c|}{\cellcolor[HTML]{C0C0C0}46.39\%}          & \multicolumn{1}{c|}{41.79\%}          & \multicolumn{1}{c|}{47.06\%}          & \multicolumn{1}{c|}{54.96\%}          & \multicolumn{1}{c|}{\hl{61.82\%}} & \multicolumn{1}{c|}{28.5718} & \multicolumn{1}{c|}{0.839}  & \multicolumn{1}{c|}{16.0097} & 0.1743             \\
\multicolumn{1}{c|}{\multirow{-2}{*}{$p_1$}}  & \multicolumn{1}{c|}{16} & \multicolumn{1}{c|}{\hl{58.76\%}} & \multicolumn{1}{c|}{\cellcolor[HTML]{C0C0C0}\hl{69.50\%}} & \multicolumn{1}{c|}{\hl{46.67\%}} & \multicolumn{1}{c|}{\hl{49.77\%}} & \multicolumn{1}{c|}{\hl{58.80\%}} & \multicolumn{1}{c|}{60.45\%}          & \multicolumn{1}{c|}{28.2141} & \multicolumn{1}{c|}{0.8194} & \multicolumn{1}{c|}{19.1739} & 0.1964             \\ \hline
\multicolumn{1}{c|}{}                         & \multicolumn{1}{c|}{8}  & \multicolumn{1}{c|}{\hl{51.13\%}} & \multicolumn{1}{c|}{\cellcolor[HTML]{C0C0C0}32.48\%}          & \multicolumn{1}{c|}{\hl{38.81\%}} & \multicolumn{1}{c|}{\hl{43.92\%}} & \multicolumn{1}{c|}{\hl{50.20\%}} & \multicolumn{1}{c|}{\hl{60.52\%}} & \multicolumn{1}{c|}{27.2604} & \multicolumn{1}{c|}{0.8046} & \multicolumn{1}{c|}{19.7477} & 0.2008             \\
\multicolumn{1}{c|}{\multirow{-2}{*}{$p_1+$}} & \multicolumn{1}{c|}{16} & \multicolumn{1}{c|}{49.68\%}          & \multicolumn{1}{c|}{\cellcolor[HTML]{C0C0C0}\hl{44.52\%}} & \multicolumn{1}{c|}{36.04\%}          & \multicolumn{1}{c|}{41.27\%}          & \multicolumn{1}{c|}{50.08\%}          & \multicolumn{1}{c|}{53.31\%}          & \multicolumn{1}{c|}{27.1135} & \multicolumn{1}{c|}{0.7968} & \multicolumn{1}{c|}{20.9897} & 0.2049             \\ \hline
\multicolumn{1}{c|}{}                         & \multicolumn{1}{c|}{8}  & \multicolumn{1}{c|}{\hl{80.49\%}} & \multicolumn{1}{c|}{\cellcolor[HTML]{C0C0C0}89.92\%}          & \multicolumn{1}{c|}{69.90\%}          & \multicolumn{1}{c|}{72.55\%}          & \multicolumn{1}{c|}{79.37\%}          & \multicolumn{1}{c|}{77.22\%}          & \multicolumn{1}{c|}{26.8757} & \multicolumn{1}{c|}{0.8189} & \multicolumn{1}{c|}{31.0186} & 0.2237             \\
\multicolumn{1}{c|}{\multirow{-2}{*}{$p_2$}}  & \multicolumn{1}{c|}{16} & \multicolumn{1}{c|}{80.25\%}          & \multicolumn{1}{c|}{\cellcolor[HTML]{C0C0C0}\hl{95.50\%}} & \multicolumn{1}{c|}{\hl{75.68\%}} & \multicolumn{1}{c|}{\hl{73.60\%}} & \multicolumn{1}{c|}{\hl{80.13\%}} & \multicolumn{1}{c|}{\hl{75.61\%}} & \multicolumn{1}{c|}{26.4921} & \multicolumn{1}{c|}{0.797}  & \multicolumn{1}{c|}{36.2468} & 0.2549             \\ \hline
\multicolumn{1}{c|}{}                         & \multicolumn{1}{c|}{8}  & \multicolumn{1}{c|}{\hl{76.39\%}} & \multicolumn{1}{c|}{\cellcolor[HTML]{C0C0C0}80.57\%}          & \multicolumn{1}{c|}{66.17\%}          & \multicolumn{1}{c|}{67.45\%}          & \multicolumn{1}{c|}{79.17\%}          & \multicolumn{1}{c|}{\hl{72.89\%}} & \multicolumn{1}{c|}{24.9906} & \multicolumn{1}{c|}{0.7804} & \multicolumn{1}{c|}{40.5354} & 0.2662             \\
\multicolumn{1}{c|}{\multirow{-2}{*}{$p_2+$}} & \multicolumn{1}{c|}{16} & \multicolumn{1}{c|}{76.27\%}          & \multicolumn{1}{c|}{\cellcolor[HTML]{C0C0C0}\hl{88.95\%}} & \multicolumn{1}{c|}{\hl{72.97\%}} & \multicolumn{1}{c|}{\hl{70.71\%}} & \multicolumn{1}{c|}{\hl{79.81\%}} & \multicolumn{1}{c|}{73.17\%}          & \multicolumn{1}{c|}{24.7144} & \multicolumn{1}{c|}{0.7628} & \multicolumn{1}{c|}{44.6224} & 0.2906             \\ \hline
\end{tabular}%
% }
\caption{The ablation study for the attack budget $\epsilon$.}
\label{tab:abl:eps}
\end{table}

\begin{table}[ht]
\centering
% \resizebox{\linewidth}{!}{%
\begin{tabular}{cccccccccccc}
\hline
\multicolumn{2}{c}{}                                                    & \multicolumn{6}{c}{neural networks}                                                                                                                                                                                                                                   & \multicolumn{4}{c}{IQAs}                                                                                       \\
\multicolumn{2}{c}{\multirow{-2}{*}{$t$}}                      & VGG19                                 & RN50                                                          & WR101                                 & DN121                                 & NNv2                                  & SNv2                                  & PSNR $\uparrow$              & SSIM $\uparrow$             & FID $\downarrow$             & LPIPS $\downarrow$ \\ \hline
\multicolumn{1}{c|}{}                         & \multicolumn{1}{c|}{8}  & \multicolumn{1}{c|}{55.99\%}          & \multicolumn{1}{c|}{\cellcolor[HTML]{C0C0C0}59.79\%}          & \multicolumn{1}{c|}{44.42\%}          & \multicolumn{1}{c|}{48.76\%}          & \multicolumn{1}{c|}{58.08\%}          & \multicolumn{1}{c|}{59.61\%}          & \multicolumn{1}{c|}{28.3865} & \multicolumn{1}{c|}{0.8297} & \multicolumn{1}{c|}{17.3677} & 0.1841             \\
\multicolumn{1}{c|}{\multirow{-2}{*}{$p_1$}}  & \multicolumn{1}{c|}{10} & \multicolumn{1}{c|}{\hl{58.76\%}} & \multicolumn{1}{c|}{\cellcolor[HTML]{C0C0C0}\hl{69.50\%}} & \multicolumn{1}{c|}{\hl{46.67\%}} & \multicolumn{1}{c|}{\hl{49.77\%}} & \multicolumn{1}{c|}{\hl{58.80\%}} & \multicolumn{1}{c|}{\hl{60.45\%}} & \multicolumn{1}{c|}{28.2141} & \multicolumn{1}{c|}{0.8194} & \multicolumn{1}{c|}{19.1739} & 0.1964             \\ \hline
\multicolumn{1}{c|}{}                         & \multicolumn{1}{c|}{8}  & \multicolumn{1}{c|}{\hl{51.23\%}} & \multicolumn{1}{c|}{\cellcolor[HTML]{C0C0C0}41.35\%}          & \multicolumn{1}{c|}{35.74\%}          & \multicolumn{1}{c|}{\hl{44.61\%}} & \multicolumn{1}{c|}{\hl{50.09\%}} & \multicolumn{1}{c|}{\hl{54.76\%}} & \multicolumn{1}{c|}{27.1838} & \multicolumn{1}{c|}{0.8007} & \multicolumn{1}{c|}{20.216}  & 0.2023             \\
\multicolumn{1}{c|}{\multirow{-2}{*}{$p_1+$}} & \multicolumn{1}{c|}{10} & \multicolumn{1}{c|}{49.68\%}          & \multicolumn{1}{c|}{\cellcolor[HTML]{C0C0C0}\hl{44.52\%}} & \multicolumn{1}{c|}{\hl{36.04\%}} & \multicolumn{1}{c|}{41.27\%}          & \multicolumn{1}{c|}{50.08\%}          & \multicolumn{1}{c|}{53.31\%}          & \multicolumn{1}{c|}{27.1135} & \multicolumn{1}{c|}{0.7968} & \multicolumn{1}{c|}{20.9897} & 0.2049             \\ \hline
\multicolumn{1}{c|}{}                         & \multicolumn{1}{c|}{8}  & \multicolumn{1}{c|}{\hl{81.16\%}} & \multicolumn{1}{c|}{\cellcolor[HTML]{C0C0C0}93.85\%}          & \multicolumn{1}{c|}{75.41\%}          & \multicolumn{1}{c|}{\hl{75.62\%}} & \multicolumn{1}{c|}{\hl{80.46\%}} & \multicolumn{1}{c|}{\hl{76.70\%}} & \multicolumn{1}{c|}{26.6618} & \multicolumn{1}{c|}{0.8066} & \multicolumn{1}{c|}{33.906}  & 0.2409             \\
\multicolumn{1}{c|}{\multirow{-2}{*}{$p_2$}}  & \multicolumn{1}{c|}{10} & \multicolumn{1}{c|}{80.25\%}          & \multicolumn{1}{c|}{\cellcolor[HTML]{C0C0C0}\hl{95.50\%}} & \multicolumn{1}{c|}{\hl{75.68\%}} & \multicolumn{1}{c|}{73.60\%}          & \multicolumn{1}{c|}{80.13\%}          & \multicolumn{1}{c|}{75.61\%}          & \multicolumn{1}{c|}{26.4921} & \multicolumn{1}{c|}{0.797}  & \multicolumn{1}{c|}{36.2468} & 0.2549             \\ \hline
\multicolumn{1}{c|}{}                         & \multicolumn{1}{c|}{8}  & \multicolumn{1}{c|}{75.53\%}          & \multicolumn{1}{c|}{\cellcolor[HTML]{C0C0C0}84.48\%}          & \multicolumn{1}{c|}{67.98\%}          & \multicolumn{1}{c|}{69.65\%}          & \multicolumn{1}{c|}{\hl{81.71\%}} & \multicolumn{1}{c|}{72.62\%}          & \multicolumn{1}{c|}{24.8471} & \multicolumn{1}{c|}{0.7705} & \multicolumn{1}{c|}{43.1882} & 0.2798             \\
\multicolumn{1}{c|}{\multirow{-2}{*}{$p_2+$}} & \multicolumn{1}{c|}{10} & \multicolumn{1}{c|}{\hl{76.27\%}} & \multicolumn{1}{c|}{\cellcolor[HTML]{C0C0C0}\hl{88.95\%}} & \multicolumn{1}{c|}{\hl{72.97\%}} & \multicolumn{1}{c|}{\hl{70.71\%}} & \multicolumn{1}{c|}{79.81\%}          & \multicolumn{1}{c|}{\hl{73.17\%}} & \multicolumn{1}{c|}{24.7144} & \multicolumn{1}{c|}{0.7628} & \multicolumn{1}{c|}{44.6224} & 0.2906             \\ \hline
\end{tabular}%
% }
\caption{The ablation study for the attack steps.}
\label{tab:abl:attT}
\end{table}

\bibliography{aaai25}

\clearpage
\section*{Ethical Statement}
Our research on adversarial attacks is not intended to compromise or undermine machine learning systems. Instead, our goal is to provide advanced robustness testing baselines for researchers, thereby contributing to the development of more secure and trustworthy AI systems. By identifying potential vulnerabilities, we aim to enhance the overall reliability and safety of AI, ultimately benefiting society through more robust and dependable artificial intelligence.
\bibliography{aaai25}

\section*{Reproducibility Checklist}
This paper:
\begin{itemize}
    \item Includes a conceptual outline and/or pseudocode description of AI methods introduced (yes)
    \item Clearly delineates statements that are opinions, hypothesis, and speculation from objective facts and results (yes)
    \item Provides well marked pedagogical references for less-familiare readers to gain background necessary to replicate the paper (yes)
\end{itemize}

Does this paper make theoretical contributions? (yes)

If yes, please complete the list below.

\begin{itemize}
    \item All assumptions and restrictions are stated clearly and formally. (yes)
    \item All novel claims are stated formally (e.g., in theorem statements). (yes)
    \item Proofs of all novel claims are included. (yes)
    \item Proof sketches or intuitions are given for complex and/or novel results. (yes)
    \item Appropriate citations to theoretical tools used are given. (yes)
    \item All theoretical claims are demonstrated empirically to hold. (no)
    \item All experimental code used to eliminate or disprove claims is included. (yes)
\end{itemize}

Does this paper rely on one or more datasets? (yes)

If yes, please complete the list below.

\begin{itemize}
    \item A motivation is given for why the experiments are conducted on the selected datasets (yes)
    \item All novel datasets introduced in this paper are included in a data appendix. (NA)
    \item All novel datasets introduced in this paper will be made publicly available upon publication of the paper with a license that allows free usage for research purposes. (NA)
    \item All datasets drawn from the existing literature (potentially including authors’ own previously published work) are accompanied by appropriate citations. (yes)
    \item All datasets drawn from the existing literature (potentially including authors’ own previously published work) are publicly available. (yes)
    \item All datasets that are not publicly available are described in detail, with explanation why publicly available alternatives are not scientifically satisficing. (NA)
\end{itemize}

Does this paper include computational experiments? (yes)

If yes, please complete the list below.

\begin{itemize}
    \item Any code required for pre-processing data is included in the appendix. (no).
    \item All source code required for conducting and analyzing the experiments is included in a code appendix. (no)
    \item All source code required for conducting and analyzing the experiments will be made publicly available upon publication of the paper with a license that allows free usage for research purposes. (yes)
    \item All source code implementing new methods have comments detailing the implementation, with references to the paper where each step comes from (yes)
    \item If an algorithm depends on randomness, then the method used for setting seeds is described in a way sufficient to allow replication of results. (yes)
    \item This paper specifies the computing infrastructure used for running experiments (hardware and software), including GPU/CPU models; amount of memory; operating system; names and versions of relevant software libraries and frameworks. (yes)
    \item This paper formally describes evaluation metrics used and explains the motivation for choosing these metrics. (yes)
    \item This paper states the number of algorithm runs used to compute each reported result. (yes)
    \item Analysis of experiments goes beyond single-dimensional summaries of performance (e.g., average; median) to include measures of variation, confidence, or other distributional information. (yes)
    \item The significance of any improvement or decrease in performance is judged using appropriate statistical tests (e.g., Wilcoxon signed-rank). (yes)
    \item This paper lists all final (hyper-)parameters used for each model/algorithm in the paper’s experiments. (yes)
    \item This paper states the number and range of values tried per (hyper-) parameter during development of the paper, along with the criterion used for selecting the final parameter setting. (yes)
\end{itemize}

\end{document}